\documentclass[lettersize,journal]{IEEEtran}
\usepackage{amsmath,amsfonts}
\usepackage{algorithmic}
\usepackage{algorithm}
\usepackage{array}
\usepackage[caption=false,font=normalsize,labelfont=sf,textfont=sf]{subfig}
\usepackage{textcomp}
\usepackage{stfloats}
\usepackage{url}
\usepackage{verbatim}
\usepackage{graphicx}
\usepackage{cite}
\usepackage{adjustbox}
\usepackage{booktabs}
\usepackage{colortbl}
\usepackage{multicol}
\usepackage{multirow}
\usepackage{xcolor}
\usepackage{hyperref}
\usepackage{cleveref}
\begin{document}

\title{Griffon-G: Bridging Vision-Language and Vision-Centric Tasks via Large Multimodal Models}

\author{Yufei Zhan, Hongyin Zhao, Yousong Zhu*, Fan Yang, Ming Tang, IEEE Member, Jinqiao Wang, IEEE Member
\thanks{Yufei Zhan, Hongyin Zhao, Yousong Zhu, Fan Yang, Ming Tang, and Jinqiao Wang are with the Foundation Model Research Center, Institute of Automation, Chinese Academy of Sciences, Beijing 100190, China. Yufei Zhan and Jinqiao Wang are also with the School of Artificial Intelligence, University of Chinese Academy of Sciences, Beijing 100049, China. Fan Yang and Jinqiao Wang are with the Peng Cheng Laboratory, Shenzhen 518066, China. Jinqiao Wang is also with Wuhan AI Research, Wuhan, China. Email: \{zhanyufei2021, yangfan\_2022, zhaohongyin2020\}$@$ia.ac.cn, \{yousong.zhu, tangm, jqwang\}$@$nlpr.ia.ac.cn.}
\thanks{Corresponding author: Yousong Zhu}
}

\markboth{Journal of \LaTeX\ Class Files,~Vol.~14, No.~8, August~2021}%
{Shell \MakeLowercase{\textit{et al.}}: A Sample Article Using IEEEtran.cls for IEEE Journals}


\maketitle

\begin{abstract}
Large Multimodal Models (LMMs) have achieved significant breakthroughs in various vision-language and vision-centric tasks based on auto-regressive modeling. However, these models typically focus on either vision-centric tasks, such as visual grounding and region description, or vision-language tasks, like image caption and multi-scenario VQAs. None of the LMMs have yet comprehensively unified both types of tasks within a single model, as seen in Large Language Models in the natural language processing field. Furthermore, even with abundant multi-task instruction-following data, directly stacking these data for universal capabilities extension remains challenging. To address these issues, we introduce a novel multi-dimension curated and consolidated multimodal dataset, named CCMD-8M, which overcomes the data barriers of unifying vision-centric and vision-language tasks through multi-level data curation and multi-task consolidation. More importantly, we present Griffon-G, a general large multimodal model that addresses both vision-centric and vision-language tasks within a single end-to-end paradigm. Griffon-G resolves the training collapse issue encountered during the joint optimization of these tasks, achieving better training efficiency. Evaluations across multimodal benchmarks, general Visual Question Answering (VQA) tasks, scene text-centric VQA tasks, document-related VQA tasks, Referring Expression Comprehension, and object detection demonstrate that Griffon-G surpasses the advanced LMMs and achieves expert-level performance in complicated vision-centric tasks.
\end{abstract}


\section{Introduction}

\IEEEPARstart{W}{ith} auto-regressive modeling, Large Language Models (LLMs) have successfully \cite{vicuna2023, touvron2023llama, devlin2018bert} unified various proxy tasks in natural language processing into a universal framework, contributing to a milestone. Trained on billions of tokens containing world knowledge, these models have developed the exceptional ability to perform causal reasoning for different tasks based on instructions and outperformed task-specific models, leading to their widespread application. Such success has also raised the trend in the computer vision field to adopt one universal model for different tasks, indicated as the Large Multimodal Models (LMMs) \cite{li2022blip, liu2024visual, instructblip, zhu2023minigpt, wang2023seggpt, jiao2024lumen, zhan2023griffon, UNINEXT}.

Supported by the scalability of the transformer\cite{vaswani2017attention}, rapid development has been witnessed in LMMs, but they have yet to achieve comprehensive unification like LLMs, which have unified nearly all tasks within a single framework. Some LMMs\cite{instructblip, liu2024visual, Qwen-VL} focus on vision-language tasks, such as Visual Question Answering (VQA) and image caption, and others\cite{kosmos-2, chen2023shikra} on vision-centric tasks, like region description, visual grounding, and object detection. As the capabilities of these models continue to develop, those with different focuses have begun to surpass expert models\cite{liu2023improvedllava, li2024monkey, guo2024llava-uhd, liu2024llavanext, liusphinx} or achieve expert-level performance\cite{you2023ferret, zhan2023griffon, zhan2024griffon} in their respective areas. Consequently, it is both timely and essential to consider how to unify these different types of tasks incorporating visual perception within a single model, taking a significant step closer to Artificial General Intelligence (AGI).

Simultaneously, behind the success of LMMs, the data play a crucial role in the development of model capabilities. Early approaches\cite{Qwen-VL, chen2024internvl} follow a data scaling trajectory, improving model performance by increasing the volume of data. This increase in data volume allows LMMs to better learn world knowledge and achieve generalization across more scenarios, thereby matching or even surpassing the performance of expert models. However, as model capabilities reach a bottleneck, the performance gains from simply stacking more data become very limited\cite{kosmos-2, li2024monkey}, insufficient to offset the increased consumption of training resources. Moreover, directly aggregating data from different tasks introduces the problem of data redundancy\cite{chen2023shikra, zhan2024griffon}. Therefore, the optimal approach to achieve further performance improvements and balance training costs when unifying vision-centric and vision-language tasks lies in enhancing the quality of the data used.
\begin{figure*}[t]
  \centering
  \includegraphics[width=\linewidth]{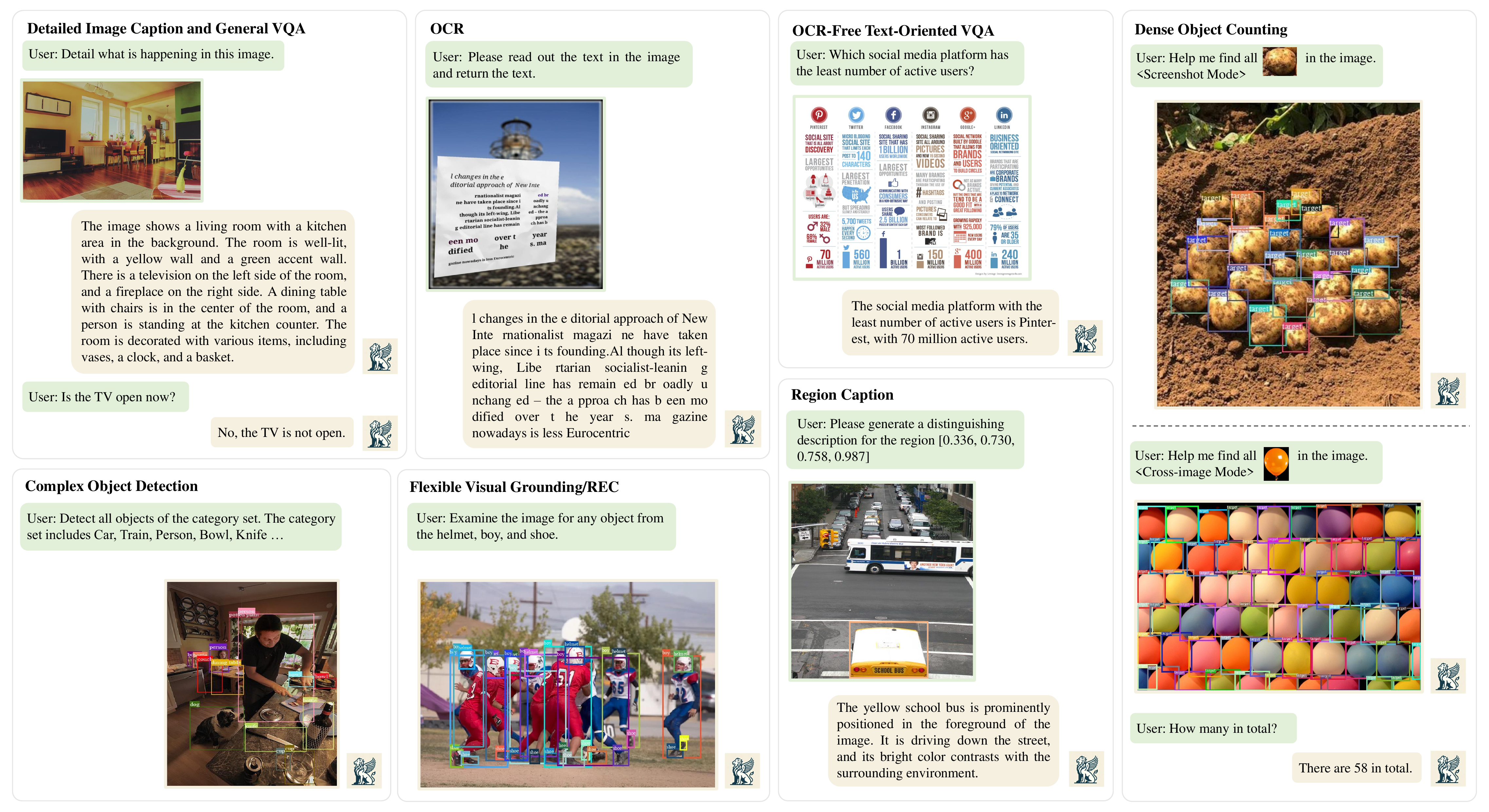}
   \caption{Overview of the Griffon-G's key capabilities. Griffon-G extends the ability boundary of general LMMs to excel in both vision-centric and vision-language tasks with one unified framework.} 
   \label{fig: page}
\end{figure*} 

In this paper, we first introduce a multi-dimension curated and consolidated multimodal dataset, named CCMD-8M. It is characterized by bridging the dataset gap between vision-centric and vision-language tasks and eliminating data redundancy from the perspectives of tasks and annotations. This dataset comprises 4.2 million curated pre-training samples and 4.1 million comprehensive instruction-following samples, which consists of ten different types of text-only, vision-centric, and vision-language tasks and serves as a critical foundation for building a general large multimodal model. Leveraging this dataset, we present Griffon-G, a unified multimodal model capable of addressing both vision-language and vision-centric tasks. More importantly, Griffon-G resolves the training collapse issue encountered during joint optimization of various tasks from different paradigms through the proposed Paradigm Progressive Learning Pipeline, which makes it possible to unify different types of tasks through auto-regressive modeling.

To further validate the capabilities of Griffon-G, we evaluate its performance on various scenarios, including the multimodal general benchmarks, general VQAs, scene text-centric VQAs, document-related VQAs, Referring Expression Comprehension (REC), and more complex object detection. Under the same amount of parameter setting, Griffon-G surpasses many recent advanced multimodal models in vision-language tasks and achieves state-of-the-art performance in tasks such as ChartQA. In vision-centric tasks, it approaches the level of advanced object detectors without relying on specialized visual expert structures and achieves state-of-the-art results in REC tasks. The contributions are summarized as follows:
\begin{itemize}
    \item We introduce a multi-dimension curated and consolidated multimodal dataset, named CCMD-8M, consisting of 4.2M pre-training data and 4.1M instruction-following data, which eliminates redundancy and improves data quality from the task and annotation perspectives.
    \item We propose the Griffon-G, a more general large vision-language model, to address both vision-language and vision-centric tasks in a unified manner. We also introduce the Paradigm Progressive Learning Pipeline to avoid the training collapse of joint optimization.
    \item We evaluate Griffon-G's performance across a wide range of benchmarks. Griffon-G achieves advanced performance in vision-language tasks and vision-centric tasks. In particular, Griffon-G outperforms classical expert models in the more challenging object detection task and achieves SOTA in the REC task, demonstrating its superior performance.
\end{itemize}

\section{Related Works}

Current LMMs can be broadly classified into two categories based on their primary focuses: vision-language models and vision-centric models.
 \begin{figure*}[t]
  \centering
  \includegraphics[width=0.8\linewidth]{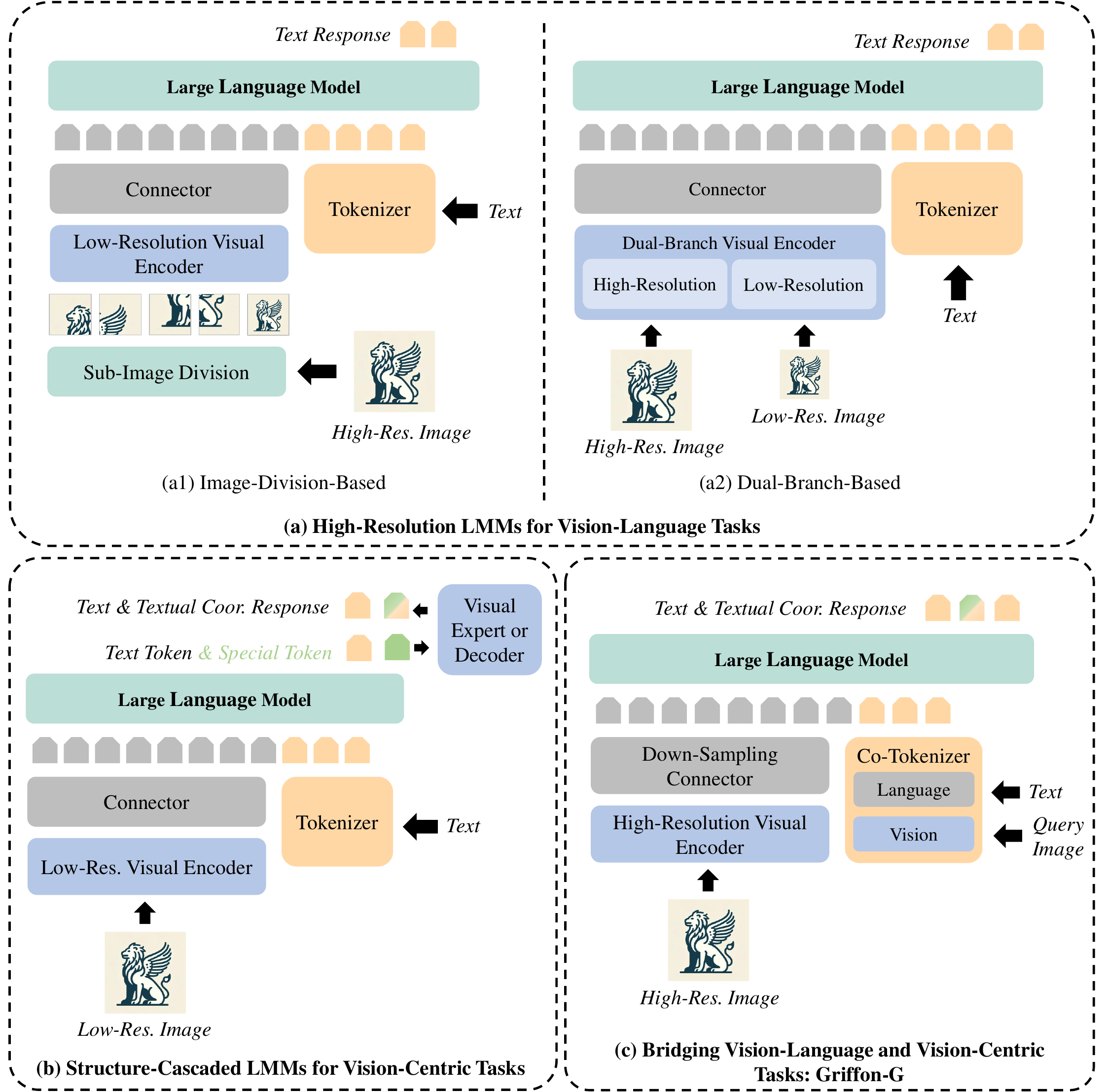}
   \caption{Structure comparison among vision-language LMMs, vision-centric LMMs, and Griffon-G. Griffon-G achieves high-resolution understanding and fine-grained localization with a streamlined high-resolution architecture without complex slicing operations and additional structures.} 
   \label{fig: structure}
\end{figure*}
\subsection{Vision-Language Large Multimodal Models}

Vision-language LMMs primarily address tasks like VQA and image caption across different scenarios, emphasizing the execution of tasks following the user's instructions. Early LMMs\cite{alayrac2022flamingo, li2023blip, li2023mimic, liu2024visual, ye2023mplug, zhu2023minigpt, instructblip} typically employ a trinary structure consisting of a visual encoder, a modality connector, and an LLM. However, they differ in their modality connectors. For instance, Flamingo\cite{alayrac2022flamingo} uses a Resampler to inject visual information into the LLM, while BLIP-2\cite{li2022blip} introduces the Q-former to extract visual features into a visual sequence for the first time. LLaVA\cite{liu2024visual} further simplifies the connector to a linear layer and enhances the model’s instruction-following capabilities through visual instruction tuning. Incorporating multi-task data\cite{instructblip, liu2023improvedllava, chen2023minigptv2} into instruction tuning enhances the model's conversational abilities, which showcases its impressive visual understanding. Subsequent works\cite{Qwen-VL, wang2023cogvlm, chen2024internvl} explore scaling the visual structure and data, further enhancing the model's performance in various vision-language tasks.

However, due to the limited visual perception capabilities of LMMs, they often fail to capture image details, resulting in the hallucination\cite{liu2023mitigating, yin2023woodpecker}. To improve the model's fine-grained perception, some fine-grained image-text data\cite{chen2023sharegpt4v, chen2024far, yuan2024osprey, shi2024umg} have been proposed or collected to replace the original image-text alignment data. These methods involve using GPT-4V\cite{openai2023gpt4} to generate detailed descriptions of images and then training a description generation model with these high-quality data to expand the data scale further. Despite having detailed image-text samples, optimizing the model to learn all the information from low-resolution images is challenging. To address this, one route\cite{hong2024cogagent, lv2023kosmos, wei2025vary} designs a dual-branch visual encoder to incorporate high-resolution information. The other route like SPHINX\cite{liusphinx} introduces the image sliding window method, which slices high-resolution images into pre-training size sub-images for separate encoding. Further research following this route\cite{li2024monkey, liu2024llavanext, liu2024textmonkey, li2024llavanext-ablations} enhances the model's resolution and designs different slicing methods to support images with any aspect ratio, preserving the original structure of the images. Although these methods have achieved expert-level performance, they are still limited to vision-language tasks. This necessitates research on how to enable models to tackle both vision-language tasks and vision-centric tasks, unlocking the further potential of LMMs.

\subsection{Vision-Centric Large Multimodal Models}

Vision-centric LMMs primarily address tasks such as visual grounding, object detection, and region description, emphasizing region-level visual perception. Pioneering works\cite{kosmos-2, chen2023shikra, chen2023minigptv2, zhang2023llavaground, zhang2023gpt4roi, guo2024regiongpt} concentrate on single-object REC and region description tasks\cite{yu2016modeling, nagaraja2016modeling}. They integrate the target to be localized or the region to be described into the task instruction and represent object locations using textual coordinates or coordinate placeholders, endowing the models with initial capabilities to handle vision-centric tasks. Subsequent research\cite{you2023ferret, Qwen-VL, wang2023cogvlm} improves the model's performance in single-object vision-centric tasks by processing and incorporating data from various visual tasks and offering more flexible reference methods. Griffon\cite{zhan2023griffon} first unifies different granularity localization tasks through next-token prediction, enabling LMMs to handle more complex visual tasks such as object detection. Griffon v2\cite{zhan2024griffon} further designs fine-grained high-resolution perception structure and visual-language co-referring methods, expanding the boundaries of vision-centric tasks in LMMs to include dense scene counting and surpassing the performance of classical expert detection models. Some studies\cite{wang2024visionllm, zhao2023bubogpt, jiao2024lumen, ma2024groma} also enhance the model's positional perception abilities in visual tasks by incorporating visual expert models or decoding structures. These methods provide valuable insights for enabling LMMs to address both visual tasks and visual-language tasks. Building on these explorations, we aim to endow LMMs with the capabilities of expert models in vision-centric tasks while ensuring excellent performance across a wide range of vision-language tasks.

\section{Methodology}
In this section, we propose a general large multimodal model, Griffon-G, capable of addressing both vision-centric tasks and vision-language tasks in a unified framework, as indicated in Figure \ref{fig: page}. Here, we first give preliminaries about the four main LMM structures in Section \ref{Pre about griffon}. Then, we introduce the proposed CCMD-8M Dataset in Section \ref{Dataset}. Moreover, we detail the Paradigm Progressive Learning Pipeline in Section \ref{training pipeline}.

\subsection{Preliminaries of Vision-Language and Vision-Centric Multimodal Structures}
\label{Pre about griffon}

Due to differences in task focuses, the architectures of various LMMs often exhibit certain variations, as shown in Figure \ref{fig: structure}. As depicted in Figure \ref{fig: structure}(a), advanced LMMs for vision-language tasks design different high-resolution structures to enhance the model's understanding of image details. For instance, models \cite{li2024monkey, liusphinx} employing image partitioning methods usually design different partitioning strategies to better accommodate varying image aspect ratios. After obtaining these sub-images at pre-trained resolutions, a scaled-down raw image is concatenated at the token or channel level to compensate for the loss of context and edge information. These small sub-images are encoded in batches and sequentially projected into the text embedding space, as demonstrated in Figure \ref{fig: structure}(a1). In contrast, dual-branch-based models simultaneously provide detailed information via high-resolution encoding and condensed global information via low-resolution branch. Input images are fed into both high-resolution and low-resolution encoders, as shown in Figure \ref{fig: structure}(a2). The fine-grained and global features are then fed directly, or after interaction, into an LLM. Although these methods have advanced the performance of LMMs in vision-language tasks, the contextual details lost due to image partitioning cannot be fully compensated by low-resolution raw images, and this also introduces additional computational overhead. The same issue arises with dual-branch structures. 

For vision-centric tasks, the development of these models follows the trajectory of LMMs. To support more vision-centric tasks, LMMs are trained to generate one or more specific special tokens tailored to the task. During the sequence decoding, these tokens are identified and then used as prompts to drive visual expert models for detecting or segmenting answer-related objects \cite{lai2024lisa, rasheed2024glamm} or to be further decoded to coordinates via visual decoding structures\cite{zhang2023nextchat}, as illustrated in the top of Figure \ref{fig: structure}(b). While this type of structure leverages the specialized visual capabilities of expert models and specifically optimized structures, it also increases the parameter burden, limits the object vocabulary into the pre-trained categories, and prevents fine-grained augmentation of LLM responses by leveraging LLM pre-training knowledge.

To achieve high-resolution image understanding while maintaining the advantages of a streamlined design, we inherit the architecture of Griffon v2\cite{zhan2024griffon}. Unlike the structures mentioned earlier, Griffon v2 enhances high-resolution perception by directly converting a low-resolution pre-trained visual encoder, \textit{i.e.} CLIP-ViT-L-14@336, into a high-resolution encoder with the resolution of 1022 using bilinear interpolation. This approach, combined with fine-tuning the visual encoder, achieves better results without the need to design complex image partitioning rules. As presented in Figure \ref{fig: structure}(c), to further improve model efficiency, a down-sampling connector is introduced to compress the input image tokens and project them into the text embedding space. Additionally, Griffon v2 follows the Griffon series' approach of converting location information into textual coordinates in the dialogue form (e.g., [x1, y1, x2, y2]) to achieve unified input-output representation via the next-token prediction, without relying on visual expert models or auxiliary structures for decoding. Moreover, a visual-language co-referring is supported to refer to an object more precisely and flexibly through an additional visual tokenizer and the extension of inputting textual object coordinates, making it more user-friendly for dialogue scenarios.

\subsection{Curated and Consolidated Multimodal Dataset}
\label{Dataset}
\begin{figure*}[t]
  \centering
  \includegraphics[width=0.95\linewidth]{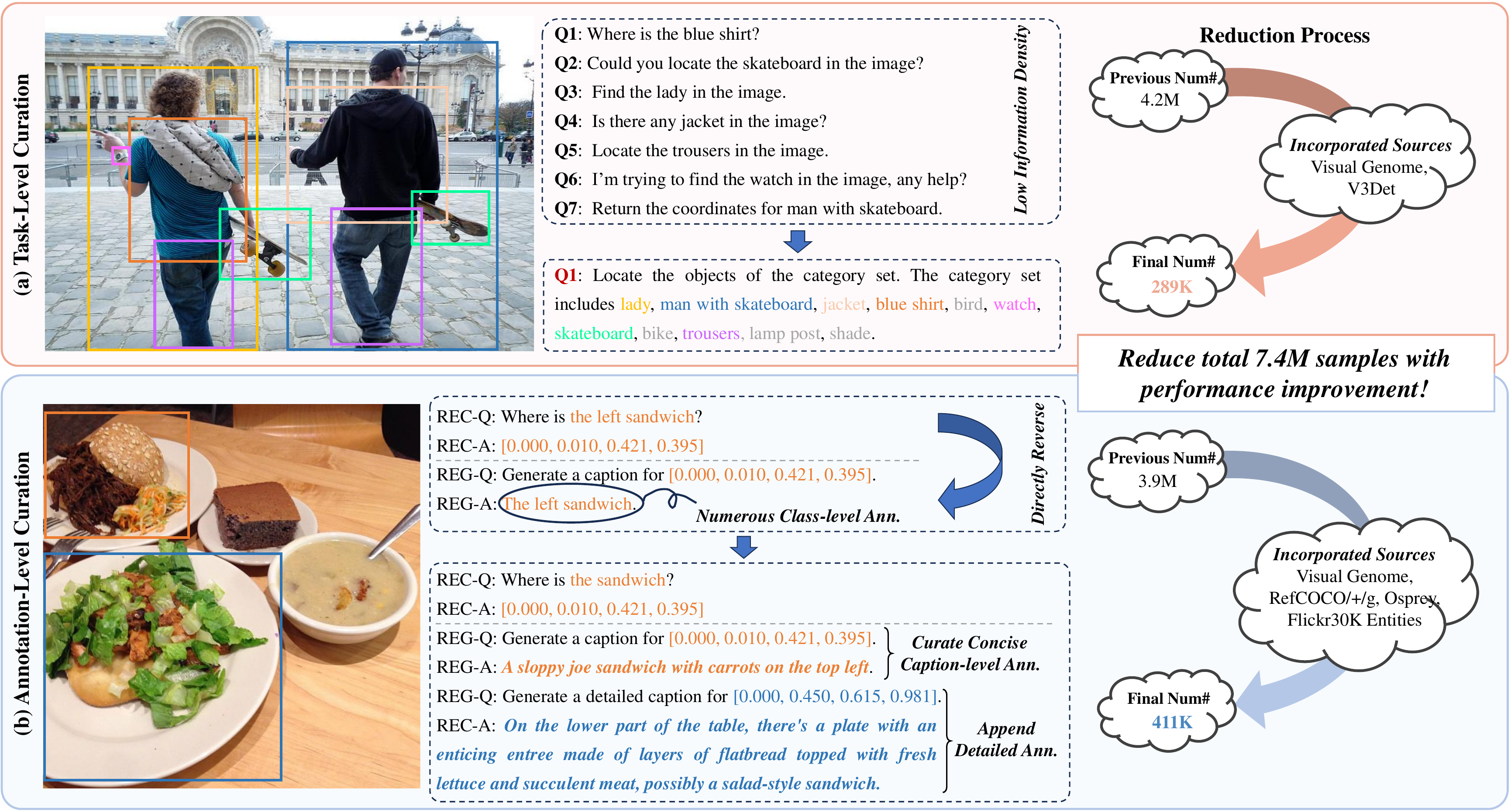}
   \caption{Illustration of the curation process of CCMD-8M Dataset. We conduct task-level curation for REC data and annotation-level curation for REG data to improve the information density, diversity, and quality.} 
   \label{fig: benchmark}
\end{figure*}
Current LMMs achieve generalization from LLMs through supervised fine-tuning with instruction data and multi-task data, commonly referred to as instruction tuning. To broaden the boundaries of model capability and enhance performance, advanced LMMs often introduce a greater variety of task types or scale-up data for each task, resulting in datasets that exceed millions of samples. Such methods are particularly prevalent in both vision-language \cite{li2024monkey, liusphinx} and vision-centric \cite{chen2023shikra, wang2023cogvlm} fields, particularly when it encompasses more challenging tasks like object detection and object counting up to ten million \cite{zhan2024griffon}. Therefore, considering the training consumption and increasing optimization challenge of over ten million data, directly combining vision-language data with vision-centric task data for joint pre-training is not suitable. Meanwhile, continuously stacking data seems less effective due to the data redundancy\cite{kosmos-2, Qwen-VL} than using well-designed data \cite{you2023ferret, liu2024llavanext, li2024llavanext-ablations}. Based on this insight, we begin by thoroughly analyzing existing vision-centric task data, and performing curation at both the task and annotation levels, resulting in a refined vision-centric set comprising 2.9M instances. Then, we consolidate the curated vision-centric data with well-designed vision-language data from multiple levels using public datasets to create the \textbf{C}urated and \textbf{C}onsolidated \textbf{M}ultimodal \textbf{D}ataset, named \textbf{CCMD-8M}. This comprehensive collection includes a 4.2M  multi-granularity pre-training set and a 4.1M diverse instruction-following set as outlined in Table \ref{tab: pre-train data} and Table \ref{tab: iinstruct data} respectively. We will elaborate this process in the following parts.

\begin{table}[t]
    \caption{Details of the CCMD-8M pre-training set with multi-level curated global alignment data and region perception data.}
    \label{tab: pre-train data}
    \centering
    \adjustbox{width=\linewidth}{%
    \begin{tabular}{c|c|c|l}
        \toprule
        Type & Format & \#samples & Data Sources \\
        \midrule
        \begin{tabular}{@{}c@{}} Global\\Alignment\end{tabular} & Caption & 1246k & ShareGPT4V\cite{chen2023sharegpt4v} \\
        \midrule
        \multirow{7}{*}{\begin{tabular}{@{}c@{}}Region\\Perception\end{tabular}} & REC       & 428K     & \begin{tabular}{@{}l@{}} RefCOCO\cite{yu2016modeling}, RefCOCO+\cite{yu2016modeling},\\RefCOCOg\cite{nagaraja2016modeling}, GRefCOCO\cite{GREC} \end{tabular}\\
        \cmidrule{2-4}
        &REG       & 411K     & \begin{tabular}{@{}l@{}l@{}} Osprey\cite{yuan2024osprey}, Flickr30K Entities\cite{flickr30k}, \\RefCOCO\cite{yu2016modeling}, RefCOCO+\cite{yu2016modeling},\\RefCOCOg\cite{nagaraja2016modeling}, Visual Genome\cite{visualgenemo}\\ \end{tabular}\\
        \cmidrule{2-4}
        &\begin{tabular}{@{}c@{}}Object\\Detection\end{tabular}       & 2107K     & \begin{tabular}{@{}l@{}} Objects365\cite{objects}, MSCOCO\cite{lin2014microsoft},\\V3Det\cite{wang2023v3det}, Visual Genome\cite{visualgenemo}\end{tabular}\\
        \midrule
        Total          & - & 4.2M & -\\
         \bottomrule
    \end{tabular}
    }
\end{table}

\textbf{Task-Level Curation.} Existing data used to build LMMs' precise visual localization capabilities for vision-centric tasks primarily consist of REC data, including datasets such as RefCOCO, Visual Genome, and GRIT, containing millions of samples. These data have endowed the LMMs with basic vision-centric task capabilities to localize the single target. However, as illustrated in Figure \ref{fig: benchmark}(a), most images contain only a single description and the corresponding object’s coordinates, resulting in relatively low information density and redundant training. Directly merging these types of data into one sample with a multi-round conversation format leads to limited task performance\cite{liu2023improvedllava}. Therefore, we propose to curate multiple vision-centric localization data into a single entry for training, thereby enhancing the information density and training efficiency of individual samples. As shown in Figure \ref{fig: benchmark}(a), we unify multiple questions of the REC task in a way akin to object detection, allowing the model to output all relevant targets at once during pre-training. Specifically, based on the visual grounding data collected in Griffon v2 \cite{zhan2024griffon}, we first curate the annotations from Visual Genome and V3Det, which occupy the largest proportion, into detection format. As indicated in the right of Figure \ref{fig: benchmark}(a), this task-level curation reduces the data volume by 3.9 million samples, improving training efficiency and enhancing model performance as discussed in Section \ref{abl}.

\textbf{Annotation-Level Curation.} Inverse to the REC task, Referring Expression Generation (REG) requires inputting the coordinates of the region to be asked and outputs the description of it. Previous methods\cite{chen2023shikra, Qwen-VL} typically use REC datasets for both REC and REG tasks or randomly sample from them to decide the form as shown in the top of Figure \ref{fig: benchmark}(b). However, this approach also leads to the low information density issue, which is exacerbated by a significant increase in data volume and quite limited performance improvement. Moreover, most of the current REG data are annotated with quite brief and close-to-category descriptions like the ``left sandwich'' exampled in Figure \ref{fig: benchmark}(b), insufficient to meet users' needs for customized regional descriptions, whether detailed or concise. Therefore, we sample the data with concise yet accurate descriptions rather than the class-level expression with the help of spaCy \cite{spaCy} which is skilled in sentence and phrase processing. As depicted in the bottom of Figure \ref{fig: benchmark}(b), the retained data align more closely with the definition of the REG task. Additionally, we augment the dataset with more detailed descriptions for the regions. For these data, we add responsive prompt words like ``more detailed'' into the instruction to differentiate and train the model to meet users' flexible needs. Specifically, we curate concise caption-level annotations from the shared Visual Genome\cite{visualgenemo} and RefCOCO/+/g\cite{yu2016modeling, nagaraja2016modeling} datasets in Griffon v2, leading to a reduction of about 3.6M samples. Moreover, we build and append detailed annotations from Osprey\cite{yuan2024osprey} and Flickr30K Entities\cite{flickr30k} with a slight data increase. With the annotation-level curation detailed above, we reduce the REG data volume from 3.9M to 411K.

Overall, through both levels of curation, we reduce the 7.4M localization-related samples from the Griffon v2 dataset to a refined vision-centric dataset containing 2.9M instances. In addition, we also add detailed vision-language alignment data from ShareGPT4V \cite{chen2023sharegpt4v} to provide a global understanding of images, and then integrate it with the curated localization data to enhance region perception and location awareness. Table \ref{tab: pre-train data} presents a detailed breakdown of our pre-training dataset.

\begin{table}[t]
    \caption{Details of the CCMD-8M instruction-following set with vision-centric and vision-language data consolidation.}
    \label{tab: iinstruct data}
    \centering
    \adjustbox{width=\linewidth}{%
    \begin{tabular}{l|c|l}
        \toprule
        Data Type & \#samples & Data Sources \\
        \midrule
        Language Instruction   & 418K     &  
        \begin{tabular}{@{}l@{}l@{}}  UltraChat\cite{ding2023enhancing}, Flan-mini\cite{ghosal2023flacuna}, OpenOrca\cite{OpenOrca},\\ ShareGPT, MetaMathQA\cite{yu2023metamath}, MathInstruct\cite{yue2023mammoth},\\ WizardCoder\cite{luo2023wizardcoder} \end{tabular}\\
        \midrule
        VL Instruction       & 1086K     & LLaVA\cite{liu2024visual}, ALLaVA\cite{chen2024allava}, LVIS-Instruct4V\cite{wang2023instruct4v}\\
        \midrule
        Image Caption       & 122K     & ShareGPT4V\cite{chen2023sharegpt4v}, TextCaps\cite{sidorov2020textcaps}\\
        \midrule
        General VQAs & 313K & \begin{tabular}{@{}l@{}}VQA v2\cite{vqa}, GQA\cite{gqa}, OK-VQA\cite{marino2019ok}, \\A-OKVQA\cite{schwenk2022okvqa}, SQA\cite{sqa}, VizWiz\cite{bigham2010vizwiz} \end{tabular}\\
        \midrule
        Scene Text-centric VQAs & 206K & \begin{tabular}{@{}l@{}}TextVQA\cite{textvqa}, OCR-VQA\cite{ocrvqa}, AI2D\cite{ai2d},\\ Synthdog\cite{kim2022donut}\end{tabular} \\
        \midrule
        Document-related VQAs & 255K  & \begin{tabular}{@{}l@{}l@{}} DVQA\cite{kafle2018dvqa}, ChartQA\cite{masry-etal-2022-chartqa}, DocVQA\cite{mathew2020docvqa},\\ InfoVQA\cite{mathew2022infographicvqa}, DeepForm\cite{borchmann2021due}, KLC\cite{borchmann2021due}, \\WTQ\cite{borchmann2021due}, TabFact\cite{borchmann2021due} \end{tabular}\\
        \midrule
        Object Detection & 463K & \begin{tabular}{@{}l@{}} Objects365\cite{objects}, MSCOCO\cite{lin2014microsoft}, V3Det\cite{wang2023v3det},\\ Visual Genome\cite{visualgenemo} \end{tabular}\\
        \midrule
        REC & 256K & RefCOCO/+\cite{yu2016modeling}, RefCOCOg\cite{nagaraja2016modeling}, GRefCOCO\cite{GREC}\\
        \midrule
        Visual Grounding & 157K & Visual Genome\cite{visualgenemo}, V3Det\cite{wang2023v3det}, Self-Collected\\
        \midrule
        REG & 171K & \begin{tabular}{@{}l@{}}Osprey\cite{yuan2024osprey}, Flickr30K Entities\cite{flickr30k}, RefCOCO/+\cite{yu2016modeling},\\ RefCOCOg\cite{nagaraja2016modeling}, Visual Genome\cite{visualgenemo} \end{tabular}\\
        \midrule
        Object Counting & 598K & Open Images\cite{OpenImages}, FSCD\cite{fscd}, Griffon v2\cite{zhan2024griffon} \\
        \midrule
        Total          & 4.1M & -\\
         \bottomrule
    \end{tabular}
    }
\end{table}

\textbf{Multi-Level Consolidation.} Current multimodal datasets adopt a paradigm where pre-training data is used for foundational capability building and instruction-following data are utilized for generalization across various tasks and scenarios. Based on the previous extensive validation\cite{liu2023improvedllava}, we employ this paradigm to construct our CCMD-8M instruction-following data. However, directly mixing different multimodal datasets may result in redundancy or omissions. To address this, we propose to conduct multi-level consolidation. We start with the core types of vision-language and vision-centric data, merging different datasets within each category according to various scenarios. In the vision-language domain, we focus on six distinct scenarios, including language and vision-language self-instruction data, general VQA, scene text-centric VQA, document-related VQA, and image caption, and gather data from various public datasets. For each scenario, we include diverse subtypes of data, such as dialogues, mathematics, and codes within language instructions, as well as PDFs and tables in document-related VQAs, and incorporate different dialogue formats like multiple-choice and open-ended questions. In the vision-centric domain, we initially collect data based on task difficulty from our curated vision-centric instances, ranging from single-object localization (REC) to dense object detection. Additionally, we reorganize existing visual grounding data to include 0-10 different category targets in each entry instead of a single category, to bridge the gap between single-object REC and dense detection while enhancing user-friendliness. Subsequently, we collect visual referring data with object counting datasets from the multi-level referring perspective and region description data with REG datasets from the task-format perspective to ensure the completeness of vision-centric data. Through these multi-level consolidations, we obtain 4.1 million instruction data, as detailed in Table \ref{tab: iinstruct data}.

\subsection{Paradigm Progressive Learning Pipeline}
\label{training pipeline}
\begin{figure*}[t]
  \centering
  \includegraphics[width=\linewidth]{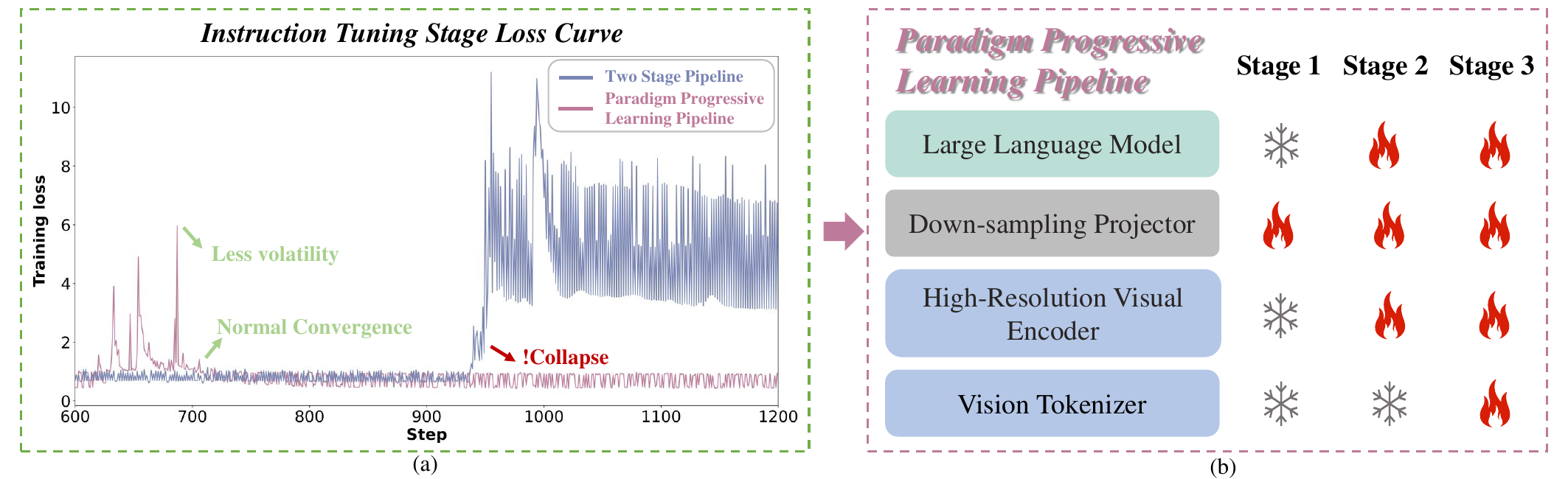}
   \caption{Visualization of Paradigm Progressive Learning Pipeline. With our proposed PPLP, the loss of the instruction tuning stage converges as normal without the training collapse when generalizing beyond vision-language tasks to vision-centric tasks.} 
   \label{fig: loss curve}
\end{figure*}

Current LMMs typically adopt a two-stage training process \cite{liu2024visual}. In the first stage, referred to as modality alignment, a learnable projector is trained on image-caption data to align the visual encoder with the LLM. In the second stage, the model undergoes supervised fine-tuning using instruction-following data across a range of tasks, enhancing its ability to generalize. This method has been widely successful across various types of LMMs, including those focused on vision-language tasks, vision-centric tasks, and image generation. To unify vision-centric and vision-language tasks within a single LMM, we initially apply the two-stage training approach using pre-training and instruction data that encompass both task types, sourced from CCMD-8M. However, as illustrated by the blue line in Figure \ref{fig: loss curve}(a), the model experiences significant fluctuations and early collapse during training. This instability arises from the granularity gap between vision-language and vision-centric tasks, a challenge that is not adequately addressed by the pre-training stage in the current two-stage pipeline.

To address this issue, we conceptualize the granularity gap between tasks as a paradigm difference and introduce our Paradigm Progressive Learning Pipeline (PPLP), inspired by human cognitive development \cite{huitt2003piaget} and also curriculum learning \cite{bengio2009curriculum}. Like children who first learn to perceive the overall scene before identifying individual objects \cite{huitt2003piaget}, we begin by training the model to align the visual encoder with the language model, enabling it to grasp basic visual content. After this alignment phase, we pre-train the model with vision-centric data to build intrinsic fine-grained visual perception. As in human cognition, where various abilities are integrated for systematic reasoning, we conduct instruction tuning with a combination of vision-language and vision-centric data from Table \ref{tab: iinstruct data} to develop robust problem-solving capabilities. The mutual reinforcement between the precise localization enabled by vision-centric data and the detailed semantic extraction from vision-language data can further enhance the model's ability to handle tasks of different granularity. As shown by the purple line in Figure \ref{fig: loss curve}(a), the proposed pipeline results in reduced volatility during training and facilitates successful model convergence. The details of the pipeline are discussed below, with a summary provided in Figure \ref{fig: loss curve}(b).

\textbf{Stage 1: Modality Alignment Initialization.} Following previous methods, we only train the randomly initialized down-sampling projector with fine-grained image-text caption data from the pre-training set of CCMD-8M as indicated by the first column of Figure \ref{fig: loss curve}(b). This stage effectively aligns the visual and language modalities and provides a solid initialization for pre-training.

\textbf{Stage 2: Paradigm Pre-Adaptation Pre-training.} To unify different types of tasks and facilitate successful convergence during the joint instruction tuning, we design this stage to equip the model with intrinsic fine-grained visual perception capabilities to mitigate the effect of paradigm differences. We pre-train the entire network except for the visual tokenizer as visualized in the second column of Figure \ref{fig: loss curve}(b) using the region perception pre-training set from the CCMD-8M dataset including multi-scene visual localization and region caption data in even complex scenarios. To preserve the language reasoning capabilities learned from large-scale LLM pre-training, we empirically replay pure text data sampled from math-related, code-related, and instruction datasets.

\textbf{Stage 3: Comprehensive Instruction Tuning.} In this stage, we use the constructed instruction-following data including both the vision-centric and vision-language data from the proposed CCMD-8M dataset for supervised fine-tuning to endow the model with systematic problem-solving capabilities benefiting from the mutual reinforcement. Specifically, we fine-tune the high-resolution visual encoder, down-sampling projector, and LLM with the instruction data, while only using the visual referring object counting data to update the visual tokenizer. By aligning the projection layer within the tokenizer in this way, we leverage the model’s learned localization capabilities for stable training.

\section{Experiments}
\label{EXP}
\subsection{Implementation Details}

\textbf{Model Setting.} In our experiment, we follow Griffon v2\cite{zhan2024griffon} to set the input resolution to 1022 and the convolution layer of the down-sampling projector with a kernel size of 3, a stride of 2, and padding of 1. We utilize the CLIP-ViT-L/14-336\cite{radford2021clip} to initialize the visual encoder with position embedding interpolation, which we provide further explanation in the ablation study, and the LLaMA2-7B/13B\cite{touvron2023llama} and Gemma2-7B/27B\cite{team2024gemma} to initialize the LLM leaving the down-sampling projector and projector of visual tokenizer random initialized. 

\textbf{Training Configuration.} We train our Griffon-G with the proposed three-stage Paradigm Progressive Learning Pipeline. We utilize the AdamW optimizer\cite{adam}, setting the learning rate to 1e-3 for the first modality alignment initialization stage and 2e-5 for the paradigm pre-adaptation pre-training stage. Also, we reduce the learning rate by half for the 27B model. For the third comprehensive instruction tuning stage, we use the same learning rate as stage 2 but set the learning rate of the visual encoder ten times less than the other parameters. The DeepSpeed stage\cite{rasley2020deepspeed} and maximum length are set to zero2 and 2048 for the first stage and zero3 and 4096 for the second and third stages, respectively. We use a cosine learning rate strategy\cite{cosine} with a warmup ratio\cite{warmup} of 0.3 and train each stage for 1 epoch with a batch size of 256.

\textbf{Evaluation.} 
To comprehensively and fairly compare our method with current advancing approaches, we conduct evaluation across six areas: Multimodal Large Language Model (MLLM) benchmarks, general VQA, scene text-centric and document-related VQA, REC, and object detection. Each of the first four areas includes multiple distinct test sets. We used widely adopted standards for evaluating each test dataset to ensure consistency with other methods, such as mAP for object detection and accuracy for various VQA tasks.
\begin{table*}[t]
\centering
\caption{Performance comparison with SOTA methods on popular MLLM benchmarks and general VQAs. - indicates the result is not reported in the original paper, while $\dagger$ indicates the ensemble of multiple visual encoders.}
\label{table:mm}
\adjustbox{width=\textwidth}{%
\setlength{\tabcolsep}{4pt}
\begin{tabular}{@{}l|c|cc|ccccc|ccccc@{}}
\toprule
Methods & LLM & Max Res. & \begin{tabular}{@{}c@{}}Precise Visual \\ Localization\end{tabular}& POPE & MME\textsuperscript{P} & MME\textsuperscript{C} & MMB  & MMMU  & VQA\textsuperscript{v2} & GQA & SQA & VizWiz  \\ 
\midrule
\multicolumn{13}{c}{\textit{Normal-Resolution Models}}\\
\midrule
BLIP-2\cite{li2023blip}           & Vicuna-13B &224 &$\times$ & 85.3   & 1294 & -     & -    & -    & 65.0    & 41.0 & -    & 19.6  \\
InstructBLIP\cite{instructblip}  & Vicuna-7B &224  &$\times$& -      & -    & -     & 36.0 & -    & -    & 49.2 & -    & 34.5  \\
InstructBLIP\cite{instructblip} & Vicuna-13B &224  &$\times$&78.9   & 1213 & -     & -    & 32.9 & -    & 49.5    & -    & 33.4    \\

Shikra\cite{chen2023shikra}           & Vicuna-13B &224 & Partial&83.9   & -    & -     & -    & -    & 77.5    & -    & -    & -    \\

Qwen-VL\cite{Qwen-VL}  & Qwen-7B &448 & Partial&  -   & 1488 & \underline{361}   & 60.6 & -    & 78.2 & 57.5 & 68.2 & 38.9 \\

LLaVA1.5\cite{liu2023improvedllava}      & Vicuna-7B&336  &$\times$&85.9   & 1511 & -     & 64.3 & 35.3 & 78.5 & 62.0 & 66.8 & 50.0 \\
LLaVA1.5\cite{liu2023improvedllava}     & Vicuna-13B &336 &$\times$&85.9   & 1531 & 295   & 67.7 & 34.8 & 80.0 & 62.9 & 69.1 & 46.8 \\
mPLUG-Owl2\cite{ye2024mplug} & LLaMA2-7B & 448 & $\times$ & - & 1450 & - & 64.5 & - & 79.4 & 56.1 & 68.7 & 54.5 \\
InternVL\cite{chen2024internvl} & Vicuna-7B & 336 & $\times$ &  86.4 & 1525 & - & - & - & 79.3 & 62.9 & - & 52.5 \\
InternVL\cite{chen2024internvl} & Vicuna-13B & 336 & $\times$ &  87.1 & 1547 & - & - & - & 80.2 & 63.9 & - & 54.6 \\

\midrule
\multicolumn{13}{c}{\textit{Image-Partition-Based High-Resolution Models}}\\
\midrule

Monkey\cite{li2024monkey}        & Qwen-7B&896 &$\times$& -      &1505  &-      & -    & -    & 80.3 & 60.7 & 69.4 & 61.2 \\
SPHINX\cite{liusphinx}  & LLaMA2-13B &672$\dagger$  & Partial& \textbf{89.1}   &1458  &284    & 71.0 & -    & -    & -    & 74.2 & -    \\
LLaVA-NeXt\cite{liu2024llavanext}   & Vicuna-7B  &672   &$\times$& 86.5   &1519  & 332   & 67.4    & 36.4 & 81.8 & 64.2 & 70.1 & 57.6 \\
LLaVA-NeXt\cite{liu2024llavanext}   & Vicuna-13B &672   &$\times$&86.2   &\textbf{1575}  &326    & 68.7    & 35.8 & 82.8 & \textbf{75.9} & 65.4 & 60.5 \\
LLaVA-UHD\cite{guo2024llava-uhd}    & Vicuna-13B &1008 &$\times$&\textbf{89.1}   &1535  &-      & 68.0 & -    & 81.7 & 65.2 & 72.0 & 56.1 \\

Ferret v2\cite{you2024ferret}   & Vicuna-7B & 1344$\dagger$ & Partial&87.8   & 1510 &-      & -    & - & 81.5 & 64.7 & -    & - \\
Ferret v2\cite{you2024ferret}   & Vicuna-13B & 1344$\dagger$ & Partial&88.1   & 1521 &-      & -    & - & 81.8 & 64.8 & -    & - \\

\midrule
\multicolumn{13}{c}{\textit{Single-Branch High-Resolution Models}}\\
\midrule

OtterHD\cite{li2023otterhd} & Persimmon-8B & 1024  &$\times$&81.5 & 1314 & 289 & 53.6 & - & - & - & - & -\\
OMG-LLaVA\cite{zhang2024omgllava} & InterLM2-7B & 1024  & Partial & 80.0 & 1177 & 235 & 47.9 & - & -& -& 57.8 & -\\
\rowcolor[gray]{0.95}
Griffon-G & LLaMA2-7B        &1022&  $\surd$ & \underline{89.0} & 1514 & 323 & 74.7 & 34.2 & 82.5 & 63.4 & 82.9 & 69.9\\
\rowcolor[gray]{0.95}
Griffon-G & LLaMA2-13B       &1022& $\surd$ &88.9   &1551  &322    &72.3  & \underline{38.4} & 83.0 & 65.4 & 85.7 & \textbf{71.2}  \\
\rowcolor[gray]{0.95}
Griffon-G & Gemma2-9B        &1022 &  $\surd$ & \textbf{89.1} & 1510 & 324 & \textbf{78.1} & 37.7 & \underline{83.7} & 65.1 & \underline{89.2} & 70.3\\
\rowcolor[gray]{0.95}
Griffon-G & Gemma2-27B        &1022 & $\surd$& \textbf{89.1} & \underline{1572} & \textbf{365} & \underline{78.0} & \textbf{45.0} & \textbf{84.2} & \underline{65.9} & \textbf{89.8} & \underline{70.7}\\
\bottomrule
\end{tabular}
} 
\vspace{-3mm}
\end{table*}

\subsection{MLLM Benchmarks and General VQAs}

We compare our method with recent LMMs on general VQA tasks and comprehensive multimodal benchmarks. In our evaluation, for general VQA tasks, we utilize VQA v2\cite{vqa}, GQA\cite{gqa}, ScienceQA\cite{sqa}, and VizWiz\cite{bigham2010vizwiz}. For comprehensive capability assessment, our evaluation includes POPE\cite{pope}, MME\cite{fu2023mme}, MMB\cite{MMBench}, and MMMU\cite{yue2023mmmu} which is a college-level multimodal comprehensive evaluation benchmark, including disciplines such as Art, Business, Health \& Medicine, Science, Humanities \& Social Science, and Tech \& Engineering, encompassing over 30 subfields.

As shown in Table \ref{table:mm}, compared to normal-resolution LMMs, both image-partition-based high-resolution models and single-branch high-resolution models like Griffon-G achieve significant improvements. Under the single-branch high-resolution architecture, our Griffon-G 7B model significantly outperforms the OtterHD and the OMG-LLaVA. When compared to the latest image-partition-based high-resolution model, Ferret v2, which has some fine-grained visual localization capabilities (such as REC), our Griffon-G model surpasses Ferret v2 in almost all evaluations at comparable parameter scales of 7B/9B and 13B. Meanwhile, Griffon-G with 27B parameters achieves state-of-the-art results in the POPE, MME cognition, and MMMU benchmarks as well as general VQA tasks including VQA and ScienceQA, and achieves comparable performance in other evaluations with second-place results. Furthermore, for models with partial precise visual localization abilities like REC, Griffion-G outperforms them under the same parameters by a large margin like the OMG-LLaVA with more comprehensive localization capabilities. These outstanding results validate the effectiveness of our entire pipeline, which successfully combines advanced vision-language task capabilities with precise vision-centric task capabilities within a single LMM.
\begin{table}[t]
\centering
\caption{{Performance on OCR-related and Doc-related VQA tasks.} }
\label{table:textvqa}
\setlength{\tabcolsep}{3pt}
\begin{adjustbox}{width=\linewidth}
\begin{tabular}{l|ccccc}
\toprule
Method    & TextVQA & DocVQA & ChartQA  & InfoVQA \\ 
\midrule
\multicolumn{5}{c}{\it Specialist models} \\
\midrule
\color{gray}Donut\cite{kim2022donut}                 &\color{gray}43.5             &  \color{gray}67.5 & \color{gray}41.8  &  \color{gray}11.6  \\
\color{gray}Pix2Struct\cite{lee2023pix2struct}           & \color{gray}-                & \color{gray}76.6  &\color{gray}58.6   &  \color{gray}40.0  \\
\color{gray}UReader-7B\cite{ye2023ureader}              & \color{gray}57.6             &  \color{gray}65.4 & \color{gray}59.3  &  \color{gray}42.2  \\
\color{gray}mPLUG-Doc-7B\cite{hu2024mplug}            & \color{gray}52.6             & \color{gray}62.2  & \color{gray}57.4  &  \color{gray}38.2 \\
\color{gray}HRVDA-7B\cite{liu2024hrvda}               & \color{gray}73.3             & \color{gray}72.1 & \color{gray}67.6     & \color{gray}43.5 \\
\color{gray}TextMonkey-7B\cite{liu2024textmonkey}           & \color{gray}64.3             & \color{gray}-    & \color{gray}58.2     & \color{gray}28.2 \\
\midrule
\multicolumn{5}{c}{\it Generalist models} \\
\midrule
Qwen-VL-7B-Chat\cite{Qwen-VL}      &  61.5             & 62.6 & 66.3  & -  \\
LLaVA1.5-7B\cite{liu2023improvedllava}      & 46.1    & 28.1    & 18.2     & 25.8\\
LLaVA1.5-13B\cite{liu2023improvedllava}     & 48.7 & 30.3    & 18.2        & 29.4  \\ 
mPLUG-Owl2-7B\cite{ye2024mplug} & 58.2 & - & - & -\\
InternVL-7B\cite{chen2024internvl} & 57.0 & - & - & - \\
InternVL-13B\cite{chen2024internvl} & 58.7 & - & - & - \\
\midrule
Vary-7B\cite{wei2025vary} & - & 76.3 & 66.1 & -\\
SPHINX-13B\cite{liusphinx}   & 65.7             & 61.2 & 53.4  & 34.7 \\
Monkey-7B\cite{li2024monkey}      & 67.6             & 66.5 & 65.1  & 36.1 \\
LLaVA-NeXt-7B\cite{liu2024llavanext} &  64.9 & 74.3 & 54.8  & 37.1 \\
LLaVA-NeXt-13B\cite{liu2024llavanext}   &67.1             & \textbf{77.5} & 60.5  & 41.3 \\
LLaVA-UHD-13B\cite{guo2024llava-uhd}      & 67.7             & -    & -      & - \\
Ferret v2-7B\cite{you2024ferret}   & 61.7 & -    & -       & - \\
Ferret v2-13B\cite{you2024ferret} & 62.2 & -    & -       & - \\
\midrule
\rowcolor[gray]{0.95}
Griffon-G-7B             & 70.0             & 71.6 & 68.4 &  45.1 \\ 
\rowcolor[gray]{0.95}
Griffon-G-13B             & \underline{70.7}             & \underline{75.5} & 71.8  & \textbf{50.4} \\
\rowcolor[gray]{0.95}
Griffon-G-9B             & \underline{70.7} & 74.2 & \underline{73.8}  & \underline{48.9}  \\
\rowcolor[gray]{0.95}
Griffon-G-27B             & \textbf{73.8} & 71.4 & \textbf{74.1}  & 48.0 \\
\bottomrule
\end{tabular}
\end{adjustbox}
\end{table}

\subsection{Scene Text-Centric and Doc-Related VQAs}

Unlike general VQA tasks, scene text-centric and document-related VQAs focus on questions about the information contained in texts, charts, and documents, requiring the model to accurately interpret text within images and reason effectively to derive the correct answers. These tasks also fall under the category of OCR-oriented VQAs. In this context, the original input images often have relatively high resolutions, with text occupying small portions of the image, making fine-grained perception and detailed reasoning more challenging for models. We assess the model's performance in scene text-centric tasks using TextVQA \cite{textvqa}, and in document-related scenarios using DocVQA \cite{mathew2020docvqa}, ChartQA \cite{masry-etal-2022-chartqa} and InfoVQA \cite{mathew2022infographicvqa}. We compare the performance of specialized expert models in the text-oriented domain with general LMMs, as shown in Table \ref{table:textvqa}. Compared to text-specialized expert models, Griffon-G significantly outperforms them, similar to the trend observed in recent LMMs. For generalist models, we categorize them into three types, following established conventions. Overall, Griffon-G surpasses models like LLaVA-NeXT, LLaVA-UHD, and Ferret v2 at comparable parameter scales. Notably, in the widely evaluated TextVQA task, Griffon-G achieves a substantial improvement, reaching the highest accuracy of 73.8. These results highlight Griffon-G’s robust local perception and strong text-oriented vision-language capabilities.

\subsection{Object Detection}
As indicated in Table \ref{table:mm}, the vast majority of LMMs lack precise visual localization capabilities with object bounding boxes output. Some of them can only deal with basic single-target grounding tasks like REC, which is quite simple among vision-centric tasks. Object detection is considered one of the most general vision-centric tasks reflecting models' vision-centric capabilities, which none of the LMMs are capable of without calling expert models with APIs or incorporating detection decoding structures during training. Therefore, we primarily compare our model with classical object detection expert models and LMM models incorporating localization decoding structures on MSCOCO val2017\cite{lin2014microsoft}. In addition to the COCO evaluation metrics, we also list model resolution and the number of training epochs for all stages, as these hyper-parameters significantly impact detection performance, providing a reference for a fair comparison. As shown in Table \ref{table:det}, despite using less data, our Griffon-G-9B acquires 41.0 mAP on COCO, outperforms Griffon v2 by 2.5 points under our design, and surpasses the Lumen model that incorporates decoding structures. Additionally, it outperforms the Faster RCNN series models\cite{ren2015faster, lin2017feature} and the DAB-DETR model\cite{liu2022dab} under comparable training epochs, further narrowing the gap with models like Pix2Seq\cite{chen2021pix2seq} that have more training epochs and higher resolutions. This solid object detection performance validates that our method successfully bridges the vision-language tasks and vision-centric tasks within a single LMM with advanced performance instead of simple function unification.

\subsection{Referring Expression Comprehension}

REC requires the model to distinguish among multiple possible objects in an image based on the referring expression and to locate the object that best matches the description. This demands a higher level of understanding of object details. With the advancement of LMMs, their excellent comprehension of textual details has led to widespread exploration of REC task in the LMM field. Following previous practice, we evaluate models on the most widely used datasets: RefCOCO\cite{yu2016modeling}, RefCOCO+\cite{yu2016modeling}, and RefCOCOg\cite{nagaraja2016modeling}. We use AP50 as the metric, where a prediction is considered accurate if its Intersection over Union (IoU) with the ground truth exceeds 0.5. We compare Griffon-G with both models at normal resolution, such as Qwen-VL\cite{Qwen-VL} and Groma\cite{ma2024groma}, and high-resolution models, including SPHINX\cite{liusphinx}, Griffon v2\cite{zhan2024griffon}, and the SOTA Ferret v2. As shown in Table \ref{table: rec}, Griffon-G achieves new state-of-the-art results in the RefCOCO series with the 27B model and even the 9B model surpasses the latest SOTA Ferret v2. These superior results further highlight the effectiveness of our CCMD-8M dataset and Paradigm Progressive Learning Pipeline, which endows the model with better instruction-following capabilities and allows it to more accurately identify the object that best matches the user's description among multiple objects. This showcases Griffon-G's convincing comprehensive precise visual localization capabilities and unification.
\begin{table}[t]
\centering
\caption{{Performance Comparison on Object Detection Task.}}
\label{table:det}
\adjustbox{width=\linewidth}{%
\setlength{\tabcolsep}{2.2pt}
\begin{tabular}{@{}l|cc|cccccc}
\toprule
Methods & Res. & \begin{tabular}{@{}c@{}}Total\\Epoch\end{tabular} & mAP & AP\textsuperscript{50} & AP\textsuperscript{75} & AP\textsuperscript{S} & AP\textsuperscript{M} & AP\textsuperscript{L}  \\ 
\midrule
\multicolumn{9}{c}{\textit{Specialist Models}}\\
\midrule

    \color{gray}Faster RCNN-C4\cite{ren2015faster}   & \color{gray}1022       & \color{gray}12     & \color{gray}35.6       & \color{gray}55.7 & \color{gray}37.8      & \color{gray}17.0      & \color{gray}40.6     & \color{gray}50.3     \\
    \color{gray}Faster RCNN-FPN\cite{lin2017feature}  & \color{gray}1022       & \color{gray}12     & \color{gray}37.9       & \color{gray}58.6 & \color{gray}40.9      & \color{gray}20.4      & \color{gray}41.1     & \color{gray}50.3     \\
    \color{gray}Pix2Seq\cite{chen2021pix2seq}          & \color{gray}1333       & \color{gray}300    & \color{gray}43.0       &\color{gray} 61.0 & \color{gray}45.6      & \color{gray}25.1      & \color{gray}46.9     & \color{gray}59.4     \\
    \color{gray}DETR\cite{carion2020end}             & \color{gray}1333       & \color{gray}500    & \color{gray}42.0       & \color{gray}62.4 & \color{gray}44.2      & \color{gray}20.5      & \color{gray}45.8     & \color{gray}61.1     \\
    \color{gray}DAB-DETR\cite{liu2022dab}         & \color{gray}1333       & \color{gray}12     & \color{gray}38.0       & \color{gray}60.3 & \color{gray}39.8      & \color{gray}19.2      & \color{gray}40.9     & \color{gray}55.4     \\
    
    \midrule
    \multicolumn{9}{c}{\textit{LMMs with Detection Heads}}\\
    \midrule
    Lumen-7B\cite{jiao2024lumen}         & 448          & -      & 33.9       & 51.2 & 34.2      & -         & -        & -        \\
    \midrule
    \multicolumn{9}{c}{\textit{LMMs with Detection Abilities}}\\
    \midrule
    Griffon v2-13B\cite{zhan2024griffon}       & 1022       & 3      & 38.5       & 54.3 & 41.2      & 19.4      & 43.2     & 57.6     \\
    \rowcolor[gray]{0.95}
    Griffon-G-7B    & 1022       & 3      & 40.2       & \textbf{57.4} & 42.8      & \underline{23.5}      & 44.6     & 58.2     \\
    \rowcolor[gray]{0.95}
    Griffon-G-13B    & 1022       & 3      & 40.1       & \underline{57.0} & 42.8      & 22.2      & 44.7     & 58.9     \\
    \rowcolor[gray]{0.95}
    Griffon-G-9B    & 1022       & 3       & \textbf{41.0}   & 56.7 & \textbf{44.0} & 23.1 & \textbf{46.2} & \underline{60.6}\\
    \rowcolor[gray]{0.95}
    Griffon-G-27B    & 1022       & 3       & \underline{40.6}   & \underline{57.0} & \underline{43.1} & \textbf{24.0} & \underline{45.8} & \textbf{60.9}\\
\bottomrule
\end{tabular}
} 
\vspace{-3mm}
\end{table}

\subsection{Ablation Studies}
\label{abl}
To further validate the effectiveness of our method, following the methodology section, we conduct detailed ablation experiments from three perspectives: structure, data, and training methods. We show the results as follows.

\begin{table}[t]
\caption{{{Performance (Top-1 Accuracy@0.5) on Referring Expression Comprehension tasks.}}}
\label{table: rec}
\adjustbox{max width=\linewidth}{%

\setlength{\tabcolsep}{2.2pt}
\begin{tabular}{@{}l|ccc|ccc|cc@{}}
\toprule

\multirow{2}{*}{Method} & \multicolumn{3}{c|}{RefCOCO} & \multicolumn{3}{c|}{RefCOCO+} & \multicolumn{2}{c}{RefCOCOg} \\
 & val & test-A & test-B & val & test-A & test-B & val-u & test-u \\

\midrule

\multicolumn{9}{c}{\it Specialist models} \\

\midrule

\multicolumn{1}{l|}{\color{gray}UNINEXT\cite{UNINEXT}} & \color{gray} 92.6 & \color{gray} 94.3 & \color{gray} 91.5 & \color{gray} 85.2 & \color{gray} 89.6 & \color{gray} 79.8 & \color{gray} 88.7 & \color{gray} 89.4 \\
\multicolumn{1}{l}{\color{gray}MDETR\cite{kamath2021mdetr}} & \color{gray} 86.8 & \color{gray} 89.6 & \color{gray} 81.4 & \color{gray} 79.5 & \color{gray} 84.1 & \color{gray} 70.6 & \color{gray} 81.6 & \color{gray} 80.9 \\
\multicolumn{1}{l|}{\color{gray}G-DINO-L\cite{liu2023grounding}} & \color{gray} 90.6 & \color{gray} 93.2 & \color{gray} 88.2 & \color{gray} 82.8 & \color{gray} 89.0 & \color{gray} 75.9 & \color{gray} 86.1 & \color{gray} 87.0 \\

\midrule
\multicolumn{9}{c}{\it Generalist models} \\
\midrule
\multicolumn{1}{l|}{OFA-L\cite{wang2022ofa}}                   & 80.0 & 83.7 & 76.4 & 68.3 & 76.0 & 61.8 & 67.6 & 67.6 \\
\multicolumn{1}{l|}{MiniGPT-v2-7B\cite{chen2023minigptv2}}      & 88.1 & 91.3 & 84.3 & 79.6 & 85.5 & 73.3 & 84.2 & 84.3 \\
\multicolumn{1}{l|}{Qwen-VL-7B\cite{Qwen-VL}}         & 88.6 & 92.3 & 84.5 & 82.8 & 88.6 & 76.8 & 86.0 & 86.3 \\
\multicolumn{1}{l|}{VistaLLM-7B\cite{pramanick2024jack}}             & 88.1 & 91.5 & 83.0 & 82.9 & 89.8 & 74.8 & 83.6 & 84.4  \\
\multicolumn{1}{l|}{Groma-7B\cite{ma2024groma}}                & 89.5 & 92.1 & 86.3 & 83.9	& 88.9 & 78.1 & 86.3 & 87.0  \\
\multicolumn{1}{l|}{Ferret v2-7B}  & 92.8 & 94.7 & 88.7 & 87.4 & \underline{92.8} & 79.3 & 89.4 & 89.3  \\
\midrule
\multicolumn{1}{l|}{Shikra-13B\cite{chen2023shikra}}              & 87.8 & 91.1 & 81.8 & 82.9 & 87.8 & 74.4 & 82.6 & 83.2 \\
\multicolumn{1}{l|}{\color{gray}CogVLM-17B\cite{wang2023cogvlm}}  & \color{gray}87.5 & \color{gray}91.8 & \color{gray}81.5 & \color{gray}92.5 & \color{gray}93.9 & \color{gray}88.7 &\color{gray}89.5 & \color{gray}90.1\\
\multicolumn{1}{l|}{Ferret-13B\cite{you2023ferret}}             & 89.5 & 92.4 & 84.4 & 82.8 & 88.1 & 75.2 & 85.8 & 86.3\\
\multicolumn{1}{l|}{Griffon v1-13B\cite{zhan2023griffon}}          & 89.4 & 92.5 & 84.6 & 83.3 & 88.4 & 76.0 & 85.2 & 86.1 \\
\multicolumn{1}{l|}{SPHINX-13B\cite{liusphinx}}              & 91.1 & 92.7 & 86.6 & 86.6 & 91.1 & 80.4 & 88.2 & 88.4 \\
\multicolumn{1}{l|}{Griffon v2-13B\cite{zhan2024griffon}}    & 89.6 & 91.8 & 86.5 & 81.9 & 85.5 & 76.2 & 85.9 & 86.0  \\
\multicolumn{1}{l|}{VistaLLM-13B\cite{pramanick2024jack}}      & 89.9 & 92.5 & 85.0 & 84.1 & 90.3 & 75.8 & 86.0 & 86.4 \\
\multicolumn{1}{l|}{Ferret v2-13B}  & 92.6 & 95.0 & 88.9 & 87.4 & 92.1 & 81.4 & \underline{89.4} & \textbf{90.0}  \\
\midrule
\rowcolor[gray]{0.95}
\multicolumn{1}{l|}{\cellcolor[gray]{0.95}Griffon-G-7B} & 92.3 & 94.5 & 88.6 & 86.4 & 91.0 & 79.7 & 88.8 & 88.6 \\
\rowcolor[gray]{0.95}
\multicolumn{1}{l|}{\cellcolor[gray]{0.95}Griffon-G-13B} & 92.1 & 94.7 & 88.7 & 86.4 & 91.5 & 80.0 & 88.1 & 88.6 \\
\rowcolor[gray]{0.95}
\multicolumn{1}{l|}{\cellcolor[gray]{0.95}Griffon-G-9B} & \underline{93.2} & \underline{95.2} & \underline{89.8} & \underline{88.1} & 92.6 & \textbf{82.4} & \underline{89.4} & \underline{89.8} \\
\rowcolor[gray]{0.95}
\multicolumn{1}{l|}{\cellcolor[gray]{0.95}Griffon-G-27B} & \textbf{93.8} & \textbf{95.5} & \textbf{90.3} & \textbf{88.8} & \textbf{93.1} & \underline{82.2} & \textbf{89.5} & \underline{89.8} \\

\bottomrule
\end{tabular}
} 
\label{table:ref}
\end{table}

\begin{table}[t]
\centering
\caption{{Comparison on Different Pre-trained Visual Encoder under High-resolution Setting.} }
\label{table:Visual Encoder}
\setlength{\tabcolsep}{6pt}
\begin{adjustbox}{}
\begin{tabular}{lc|cc}
\toprule
Type & Model & TextVQA & mAP \\
\midrule
Downstream Task Pre-training      & SAM   & 44.8  & 26.5 \\
Self-Supervised Learning         & DINOv2  & 47.5  &  30.6\\
Image-Text Contrastive Learning  & CLIP  & 64.4  & 28.2 \\
\bottomrule
\end{tabular}
\end{adjustbox}
\end{table}
\textbf{Different Pre-trained Visual Encoders at High-resolution.} 
Previous methods\cite{you2024ferret} have compared different pre-trained visual encoders either under low-resolution configurations or from the perspective of LMMs' vision-language capabilities. In these comparisons, visual encoders are usually kept frozen without updates. However, with the improvement of training methods, unfreezing and fine-tuning visual encoders\cite{liu2024llavanext} has become increasingly common to achieve better results. Thus, we compare three pre-trained visual encoders under the single branch high-resolution structure from both vision-language and vision-centric perspectives, including SAM\cite{kirillov2023segany} with downstream task pre-training, DINO\cite{oquab2023dinov2} with self-supervised learning, and CLIP\cite{radford2021clip} with image-text contrastive learning. We use the TextVQA and the MSCOCO object detection to indicate the performance respectively. We use the same model size of Large and the nearly same resolutions: 1022 for DINO and CLIP, and 1024 for SAM, due to its patch size of 16. We uniformly sample a subset from our dataset for rapid comparisons. As shown in Table \ref{table:Visual Encoder}, the CLIP pre-trained model significantly outperforms others in TextVQA, achieving a margin of nearly 20\% over its closest competitor, and secures second place in object detection for visual localization. In contrast, the DINOv2 model excels in vision-centric tasks but underperforms in the vision-language task TextVQA, which is consistent with its behavior in other downstream tasks. We attribute CLIP's superior performance to the higher quality of image-text data used during pre-training and its ability to effectively align image details with text through high-resolution scaling after image-text contrastive pre-training. This fine-tuning significantly enhances its performance on fine-grained understanding tasks. Given CLIP’s stronger overall capabilities, we select it for initialization.

\textbf{Effects of Paradigm Progressive Learning Pipeline.}
As analyzed in the previous sections, due to the paradigm difference between different types of tasks, the model collapses if following the two-stage training process for task unification and struggles to converge to similar levels, leading to a decrease in performance. To address this issue, we propose the Paradigm Progressive Learning Pipeline, where the model is pre-trained to cultivate intrinsic fine-grained visual perception capabilities. As shown in Figure \ref{fig: loss curve}(a), with our pipeline the model no longer collapses and quickly converges after fluctuations, indicating the success of our pipeline qualitatively. We provide quantitative proof of our design here. As shown in Table \ref{table: PPaTP}, the initial model, which suffers from training collapses, shows significantly reduced performance in fine-grained vision-language tasks such as TextVQA and comprehensive benchmarks like MME, with its object detection performance dropping to less than 50\% its original level. In contrast, our method enables the model to excel in both vision-language and vision-centric tasks under fine-grained perception, demonstrating its superior capabilities.

\begin{table}[t]
\centering
\caption{{Ablations on the Effects of the proposed Paradigm Pre-Adaptation Training Pipeline.} }
\label{table: PPaTP}
\footnotesize
\setlength{\tabcolsep}{5pt}
\begin{adjustbox}{}
\begin{tabular}{l|cccc}
\toprule
Method & TextVQA & MME\textsuperscript{P} & MME\textsuperscript{C} & mAP \\
\midrule
Ours                                   & 70.7  & 1551 & 322 & 39.8 \\
w/o PPLP   & 63.2  & 1536 & 295 & 15.5 \\
\bottomrule
\end{tabular}
\end{adjustbox}
\end{table}

 \begin{figure*}[t]
  \centering
  \includegraphics[width=0.9\linewidth]{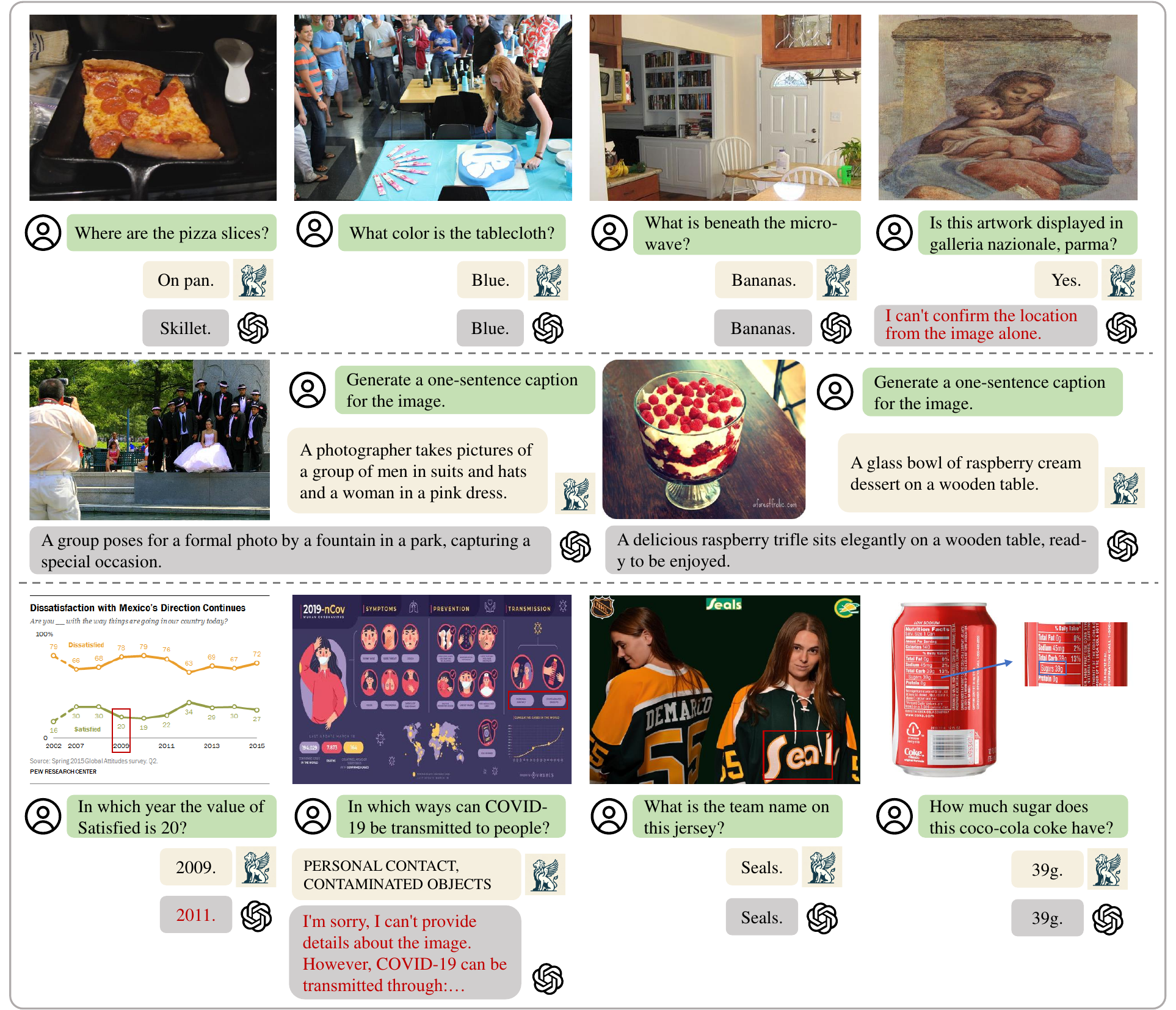}
   \caption{Qualitative analysis with more test samples on general scenarios and scenarios with dense text. Griffon-G has demonstrated impressive performance when compared to GPT-4o. The red color indicates the wrong responses.} 
   \label{fig: quant}
\end{figure*}

\textbf{Effects of Multi-Dimension Curation.} 
Multi-dimension curation plays an important role in the construction of datasets, with which we effectively reduce the volume of current large-scale vision-centric task data. Utilizing these high-quality curated visual scenario datasets for pre-training and instruction tuning, our model achieves superior vision-centric task capabilities compared to current advanced models. Using the localization data collected in Griffon v2 as our baseline, we progressively apply annotation-level curation and task-level curation. We then train models with the curated datasets at different levels to verify the effectiveness of both curation methods. We use REC and object detection tasks for evaluation, which more intuitively demonstrate the model’s visual task capabilities. Through annotation-level curation, we introduce more image sources with region descriptions while eliminating low-quality descriptions, instead of directly using REC data. This approach not only reduces training time but also avoids repetitive localization and descriptive questions for the same region in an image, significantly improving the model's performance on REC by an average of 6\%. Further task-level curation, leveraging higher-information content data and a small amount of task-specific data, further enhances the model's performance, achieving 39.0 mAP and an average REC performance of 90.0, while also improving training efficiency. These results validate the effectiveness of our designed curation method.

\begin{table}
\centering
\caption{{Ablations on the Effects of the Multi-Dimension Curation of the Dataset.} }
\label{table:curation}
\footnotesize
\setlength{\tabcolsep}{3pt}
\begin{adjustbox}{}
\begin{tabular}{l|cc}
\toprule
Method & 
\begin{tabular}{@{}c@{}}RefCOCO series\\AP\textsuperscript{50} Avg.\\ \end{tabular} & 
\begin{tabular}{@{}c@{}}COCO2017\\mAP\\ \end{tabular} \\
\midrule
Baseline                       & 78.0 & 35.9 \\
\;\; + annotation-curation     & 84.2 & 37.5 \\
\;\; + task-curation           & 90.0 & 39.0  \\
\bottomrule
\end{tabular}
\end{adjustbox}
\end{table}

\textbf{Mutual Effects between Vision-Centric and Vision-Language Tasks.}
 With the CCMD-8M dataset and our designed Paradigm Progressive Learning Pipeline, our Griffon-G bridges the fine-grained image understanding capabilities for vision-language tasks and precise visual localization capabilities for vision-centric tasks within a single LMM. Previous models lack such capabilities, and thus, there has been no analysis of how both of them impact the other one. To investigate this, we train our model by combining it with different datasets to equip it with the ability to solve corresponding tasks for comparison. Starting with instruction tuning using the LLaVA dataset\cite{liu2024visual}, we observe significant improvements across multiple tasks with our improved consolidated vision-language instructions. This is particularly evident in text-related scenarios, such as a 6.3\% increase in TextVQA performance without OCR hints. Subsequently, we train the model with the complete set of instruction data. As shown in Table \ref{table: data}, by equipping the model with precise visual reasoning capabilities even for complex scenarios, the model’s overall understanding of spatial and local details improves, thereby enhancing its vision-language task abilities and performance in corresponding tasks. Meanwhile, the precise visual localization capabilities for vision-centric tasks are also enhanced with an increase of 0.8 mAP in MSCOCO with the incorporation of vision-language tasks. These results validate the positive mutual influence between vision-centric and vision-language tasks.

\begin{table}
\centering
\caption{{Ablations on Mutual Effects between Reasoning and Localization.} ${\dagger}$ indicates using the OCR tokens as hints during evaluation.}
\label{table: data}
\footnotesize
\setlength{\tabcolsep}{3pt}
\begin{adjustbox}{}
\begin{tabular}{cc|ccc}
\toprule
\multicolumn{2}{c|}{CCMD-8M Instruction Data} & \multirow{2}{*}{VQA\textsuperscript{v2}} & \multirow{2}{*}{TextVQA} & \multirow{2}{*}{mAP} \\
Vision-Language & Vision-Centric & &  & \\
\midrule
\multicolumn{2}{c|}{LLaVA Baseline}& 81.7 & 62.8$^{\dagger}$   & - \\
$\surd$&$\times$& 82.1 & 69.1   & - \\
$\times$&$\surd$& - & - & 39.0\\
$\surd$&$\surd$& 83.0 & 70.7 & 39.8 \\
\bottomrule
\end{tabular}
\end{adjustbox}
\end{table}

\subsection{Qualitative Analysis}
We also conduct a qualitative analysis with GPT-4o among the general scenarios and text-oriented scenarios. As shown in Figure \ref{fig: quant}, Griffon-G can accurately answer the question in general scenarios and caption the image without ignoring the key details like GPT-4o. In text-oriented scenarios, Griffon-G even outperforms GPT-4o in responding to the users' questions according to the details in the image, while GPT-4o sometimes generates wrong answers.

\section{Conclusion}

In this paper, we aim to provide a more general LMM to unify vision-centric and vision-language tasks, which may broaden the applications of LMMs. To achieve this, we first introduce the CCMD-8M dataset, which addresses the issue of large volumes of low information-density vision-centric task data impeding efficient model learning and consolidates vision-centric and vision-language tasks data, to support the overall training. With the CCMD-8M dataset, we present Griffon-G, an LMM that boasts strong vision-language task abilities and precise vision-centric task capabilities. Griffon-G inherits the high-resolution structure of Griffon v2, endowing the model with complete-context fine-grained perceptual abilities. Furthermore, it addresses the training collapse issue caused by paradigm differences between different tasks which here specifically the vision-centric tasks and vision-language tasks during instruction tuning through the proposed Paradigm Progressive Learning Pipeline. We conduct comprehensive evaluations of Griffon-G on general VQAs, multimodal benchmarks, OCR-related VQA, Doc-related VQA, object detection, and REC tasks. The results demonstrate that our model surpasses the current advanced multimodal models on the most of test sets. We hope Griffon-G will serve as a foundation for the community to build more comprehensive multimodal models. 

\bibliographystyle{IEEEtran}
\bibliography{egbib}

\begin{thebibliography}{100}
\providecommand{\url}[1]{#1}
\csname url@samestyle\endcsname
\providecommand{\newblock}{\relax}
\providecommand{\bibinfo}[2]{#2}
\providecommand{\BIBentrySTDinterwordspacing}{\spaceskip=0pt\relax}
\providecommand{\BIBentryALTinterwordstretchfactor}{4}
\providecommand{\BIBentryALTinterwordspacing}{\spaceskip=\fontdimen2\font plus
\BIBentryALTinterwordstretchfactor\fontdimen3\font minus \fontdimen4\font\relax}
\providecommand{\BIBforeignlanguage}[2]{{%
\expandafter\ifx\csname l@#1\endcsname\relax
\typeout{** WARNING: IEEEtran.bst: No hyphenation pattern has been}%
\typeout{** loaded for the language `#1'. Using the pattern for}%
\typeout{** the default language instead.}%
\else
\language=\csname l@#1\endcsname
\fi
#2}}
\providecommand{\BIBdecl}{\relax}
\BIBdecl

\bibitem{vicuna2023}
\BIBentryALTinterwordspacing
W.-L. Chiang, Z.~Li, Z.~Lin, Y.~Sheng, Z.~Wu, H.~Zhang, L.~Zheng, S.~Zhuang, Y.~Zhuang, J.~E. Gonzalez, I.~Stoica, and E.~P. Xing, ``Vicuna: An open-source chatbot impressing gpt-4 with 90\%* chatgpt quality,'' March 2023. [Online]. Available: \url{https://lmsys.org/blog/2023-03-30-vicuna/}
\BIBentrySTDinterwordspacing

\bibitem{touvron2023llama}
H.~Touvron, L.~Martin, K.~Stone, P.~Albert, A.~Almahairi, Y.~Babaei, N.~Bashlykov, S.~Batra, P.~Bhargava, S.~Bhosale \emph{et~al.}, ``Llama 2: Open foundation and fine-tuned chat models,'' \emph{arXiv preprint arXiv:2307.09288}, 2023.

\bibitem{devlin2018bert}
J.~Devlin, M.-W. Chang, K.~Lee, and K.~Toutanova, ``Bert: Pre-training of deep bidirectional transformers for language understanding,'' \emph{arXiv preprint arXiv:1810.04805}, 2018.

\bibitem{li2022blip}
J.~Li, D.~Li, C.~Xiong, and S.~Hoi, ``Blip: Bootstrapping language-image pre-training for unified vision-language understanding and generation,'' in \emph{International conference on machine learning}.\hskip 1em plus 0.5em minus 0.4em\relax PMLR, 2022, pp. 12\,888--12\,900.

\bibitem{liu2024visual}
H.~Liu, C.~Li, Q.~Wu, and Y.~J. Lee, ``Visual instruction tuning,'' \emph{Advances in neural information processing systems}, vol.~36, 2024.

\bibitem{instructblip}
W.~Dai, J.~Li, D.~Li, A.~M.~H. Tiong, J.~Zhao, W.~Wang, B.~Li, P.~Fung, and S.~Hoi, ``Instructblip: Towards general-purpose vision-language models with instruction tuning,'' 2023.

\bibitem{zhu2023minigpt}
D.~Zhu, J.~Chen, X.~Shen, X.~Li, and M.~Elhoseiny, ``Minigpt-4: Enhancing vision-language understanding with advanced large language models,'' \emph{arXiv preprint arXiv:2304.10592}, 2023.

\bibitem{wang2023seggpt}
X.~Wang, X.~Zhang, Y.~Cao, W.~Wang, C.~Shen, and T.~Huang, ``Seggpt: Towards segmenting everything in context,'' in \emph{Proceedings of the IEEE/CVF International Conference on Computer Vision}, 2023, pp. 1130--1140.

\bibitem{jiao2024lumen}
Y.~Jiao, S.~Chen, Z.~Jie, J.~Chen, L.~Ma, and Y.-G. Jiang, ``Lumen: Unleashing versatile vision-centric capabilities of large multimodal models,'' \emph{arXiv preprint arXiv:2403.07304}, 2024.

\bibitem{zhan2023griffon}
Y.~Zhan, Y.~Zhu, Z.~Chen, F.~Yang, M.~Tang, and J.~Wang, ``Griffon: Spelling out all object locations at any granularity with large language models,'' in \emph{ECCV}, 2024.

\bibitem{UNINEXT}
B.~Yan, Y.~Jiang, J.~Wu, D.~Wang, Z.~Yuan, P.~Luo, and H.~Lu, ``Universal instance perception as object discovery and retrieval,'' in \emph{CVPR}, 2023.

\bibitem{vaswani2017attention}
A.~Vaswani, N.~Shazeer, N.~Parmar, J.~Uszkoreit, L.~Jones, A.~N. Gomez, {\L}.~Kaiser, and I.~Polosukhin, ``Attention is all you need,'' \emph{Advances in neural information processing systems}, vol.~30, 2017.

\bibitem{Qwen-VL}
J.~Bai and et~al, ``Qwen-vl: A versatile vision-language model for understanding, localization, text reading, and beyond,'' \emph{arXiv:2308.12966}, 2023.

\bibitem{kosmos-2}
Z.~Peng and et~al, ``Kosmos-2: Grounding multimodal large language models to the world,'' in \emph{ICLR}, 2024.

\bibitem{chen2023shikra}
K.~Chen and et~al, ``Shikra:unleashing multimodal llm's referential dialogue magic,'' \emph{arXiv:2306.15195}, 2023.

\bibitem{liu2023improvedllava}
H.~Liu, C.~Li, Y.~Li, and Y.~J. Lee, ``Improved baselines with visual instruction tuning,'' 2023.

\bibitem{li2024monkey}
Z.~Li, B.~Yang, Q.~Liu, Z.~Ma, S.~Zhang, J.~Yang, Y.~Sun, Y.~Liu, and X.~Bai, ``Monkey: Image resolution and text label are important things for large multi-modal models,'' in \emph{Proceedings of the IEEE/CVF Conference on Computer Vision and Pattern Recognition}, 2024, pp. 26\,763--26\,773.

\bibitem{guo2024llava-uhd}
Z.~Guo, R.~Xu, Y.~Yao, J.~Cui, Z.~Ni, C.~Ge, T.-S. Chua, Z.~Liu, and G.~Huang, ``{LLaVA-UHD}: an lmm perceiving any aspect ratio and high-resolution images,'' in \emph{ECCV}, 2024.

\bibitem{liu2024llavanext}
\BIBentryALTinterwordspacing
H.~Liu, C.~Li, Y.~Li, B.~Li, Y.~Zhang, S.~Shen, and Y.~J. Lee, ``Llava-next: Improved reasoning, ocr, and world knowledge,'' January 2024. [Online]. Available: \url{https://llava-vl.github.io/blog/2024-01-30-llava-next/}
\BIBentrySTDinterwordspacing

\bibitem{liusphinx}
D.~Liu, R.~Zhang, L.~Qiu, S.~Huang, W.~Lin, S.~Zhao, S.~Geng, Z.~Lin, P.~Jin, K.~Zhang \emph{et~al.}, ``Sphinx-x: Scaling data and parameters for a family of multi-modal large language models,'' in \emph{Forty-first International Conference on Machine Learning}.

\bibitem{you2023ferret}
H.~You and et~al, ``Ferret: Refer and ground anything anywhere at any granularity,'' in \emph{ICLR}, 2024.

\bibitem{zhan2024griffon}
Y.~Zhan, Y.~Zhu, H.~Zhao, F.~Yang, M.~Tang, and J.~Wang, ``Griffon v2: Advancing multimodal perception with high-resolution scaling and visual-language co-referring,'' \emph{arXiv preprint arXiv:2403.09333}, 2024.

\bibitem{chen2024internvl}
Z.~Chen, J.~Wu, W.~Wang, W.~Su, G.~Chen, S.~Xing, M.~Zhong, Q.~Zhang, X.~Zhu, L.~Lu \emph{et~al.}, ``Internvl: Scaling up vision foundation models and aligning for generic visual-linguistic tasks,'' in \emph{Proceedings of the IEEE/CVF Conference on Computer Vision and Pattern Recognition}, 2024, pp. 24\,185--24\,198.

\bibitem{alayrac2022flamingo}
J.-B. Alayrac, J.~Donahue, P.~Luc, A.~Miech, I.~Barr, Y.~Hasson, K.~Lenc, A.~Mensch, K.~Millican, M.~Reynolds \emph{et~al.}, ``Flamingo: a visual language model for few-shot learning,'' \emph{Advances in neural information processing systems}, vol.~35, pp. 23\,716--23\,736, 2022.

\bibitem{li2023blip}
J.~Li, D.~Li, S.~Savarese, and S.~Hoi, ``Blip-2: Bootstrapping language-image pre-training with frozen image encoders and large language models,'' in \emph{International conference on machine learning}.\hskip 1em plus 0.5em minus 0.4em\relax PMLR, 2023, pp. 19\,730--19\,742.

\bibitem{li2023mimic}
B.~Li, Y.~Zhang, L.~Chen, J.~Wang, F.~Pu, J.~Yang, C.~Li, and Z.~Liu, ``Mimic-it: Multi-modal in-context instruction tuning,'' \emph{arXiv preprint arXiv:2306.05425}, 2023.

\bibitem{ye2023mplug}
Q.~Ye, H.~Xu, G.~Xu, J.~Ye, M.~Yan, Y.~Zhou, J.~Wang, A.~Hu, P.~Shi, Y.~Shi \emph{et~al.}, ``mplug-owl: Modularization empowers large language models with multimodality,'' \emph{arXiv preprint arXiv:2304.14178}, 2023.

\bibitem{chen2023minigptv2}
J.~Chen, D.~Zhu, X.~Shen, X.~Li, Z.~Liu, P.~Zhang, R.~Krishnamoorthi, V.~Chandra, Y.~Xiong, and M.~Elhoseiny, ``Minigpt-v2: large language model as a unified interface for vision-language multi-task learning,'' \emph{arXiv preprint arXiv:2310.09478}, 2023.

\bibitem{wang2023cogvlm}
W.~Wang, Q.~Lv, W.~Yu, W.~Hong, J.~Qi, Y.~Wang, J.~Ji, Z.~Yang, L.~Zhao, X.~Song \emph{et~al.}, ``Cogvlm: Visual expert for pretrained language models,'' \emph{arXiv preprint arXiv:2311.03079}, 2023.

\bibitem{liu2023mitigating}
F.~Liu, K.~Lin, L.~Li, J.~Wang, Y.~Yacoob, and L.~Wang, ``Mitigating hallucination in large multi-modal models via robust instruction tuning,'' 2023.

\bibitem{yin2023woodpecker}
Y.~Shukang, F.~Chaoyou, Z.~Sirui, X.~Tong, W.~Hao, S.~Dianbo, S.~Yunhang, L.~Ke, S.~Xing, and C.~Enhong, ``Woodpecker: Hallucination correction for multimodal large language models,'' \emph{arXiv preprint arXiv:2310.16045}, 2023.

\bibitem{chen2023sharegpt4v}
L.~Chen, J.~Li, X.~Dong, P.~Zhang, C.~He, J.~Wang, F.~Zhao, and D.~Lin, ``Sharegpt4v: Improving large multi-modal models with better captions,'' \emph{arXiv preprint arXiv:2311.12793}, 2023.

\bibitem{chen2024far}
Z.~Chen, W.~Wang, H.~Tian, S.~Ye, Z.~Gao, E.~Cui, W.~Tong, K.~Hu, J.~Luo, Z.~Ma \emph{et~al.}, ``How far are we to gpt-4v? closing the gap to commercial multimodal models with open-source suites,'' \emph{arXiv preprint arXiv:2404.16821}, 2024.

\bibitem{yuan2024osprey}
Y.~Yuan, W.~Li, J.~Liu, D.~Tang, X.~Luo, C.~Qin, L.~Zhang, and J.~Zhu, ``Osprey: Pixel understanding with visual instruction tuning,'' in \emph{Proceedings of the IEEE/CVF Conference on Computer Vision and Pattern Recognition}, 2024, pp. 28\,202--28\,211.

\bibitem{shi2024umg}
B.~Shi, P.~Zhao, Z.~Wang, Y.~Zhang, Y.~Wang, J.~Li, W.~Dai, J.~Zou, H.~Xiong, Q.~Tian \emph{et~al.}, ``Umg-clip: A unified multi-granularity vision generalist for open-world understanding,'' \emph{arXiv preprint arXiv:2401.06397}, 2024.

\bibitem{openai2023gpt4}
OpenAI, ``Gpt-4 technical report,'' 2023.

\bibitem{hong2024cogagent}
W.~Hong, W.~Wang, Q.~Lv, J.~Xu, W.~Yu, J.~Ji, Y.~Wang, Z.~Wang, Y.~Dong, M.~Ding \emph{et~al.}, ``Cogagent: A visual language model for gui agents,'' in \emph{Proceedings of the IEEE/CVF Conference on Computer Vision and Pattern Recognition}, 2024, pp. 14\,281--14\,290.

\bibitem{lv2023kosmos}
T.~Lv, Y.~Huang, J.~Chen, L.~Cui, S.~Ma, Y.~Chang, S.~Huang, W.~Wang, L.~Dong, W.~Luo \emph{et~al.}, ``Kosmos-2.5: A multimodal literate model,'' \emph{arXiv preprint arXiv:2309.11419}, 2023.

\bibitem{wei2025vary}
H.~Wei, L.~Kong, J.~Chen, L.~Zhao, Z.~Ge, J.~Yang, J.~Sun, C.~Han, and X.~Zhang, ``Vary: Scaling up the vision vocabulary for large vision-language model,'' in \emph{European Conference on Computer Vision}.\hskip 1em plus 0.5em minus 0.4em\relax Springer, 2025, pp. 408--424.

\bibitem{liu2024textmonkey}
Y.~Liu, B.~Yang, Q.~Liu, Z.~Li, Z.~Ma, S.~Zhang, and X.~Bai, ``Textmonkey: An ocr-free large multimodal model for understanding document,'' \emph{arXiv preprint arXiv:2403.04473}, 2024.

\bibitem{li2024llavanext-ablations}
\BIBentryALTinterwordspacing
B.~Li, H.~Zhang, K.~Zhang, D.~Guo, Y.~Zhang, R.~Zhang, F.~Li, Z.~Liu, and C.~Li, ``Llava-next: What else influences visual instruction tuning beyond data?'' May 2024. [Online]. Available: \url{https://llava-vl.github.io/blog/2024-05-25-llava-next-ablations/}
\BIBentrySTDinterwordspacing

\bibitem{zhang2023llavaground}
H.~Zhang, H.~Li, F.~Li, T.~Ren, X.~Zou, S.~Liu, S.~Huang, J.~Gao, L.~Zhang, C.~Li \emph{et~al.}, ``Llava-grounding: Grounded visual chat with large multimodal models,'' \emph{arXiv preprint arXiv:2312.02949}, 2023.

\bibitem{zhang2023gpt4roi}
S.~Zhang, P.~Sun, S.~Chen, M.~Xiao, W.~Shao, W.~Zhang, K.~Chen, and P.~Luo, ``Gpt4roi: Instruction tuning large language model on region-of-interest,'' \emph{arXiv preprint arXiv:2307.03601}, 2023.

\bibitem{guo2024regiongpt}
Q.~Guo, S.~De~Mello, H.~Yin, W.~Byeon, K.~C. Cheung, Y.~Yu, P.~Luo, and S.~Liu, ``Regiongpt: Towards region understanding vision language model,'' in \emph{Proceedings of the IEEE/CVF Conference on Computer Vision and Pattern Recognition}, 2024, pp. 13\,796--13\,806.

\bibitem{yu2016modeling}
L.~Yu, P.~Poirson, S.~Yang, A.~C. Berg, and T.~L. Berg, ``Modeling context in referring expressions,'' in \emph{Computer Vision--ECCV 2016: 14th European Conference, Amsterdam, The Netherlands, October 11-14, 2016, Proceedings, Part II 14}.\hskip 1em plus 0.5em minus 0.4em\relax Springer, 2016, pp. 69--85.

\bibitem{nagaraja2016modeling}
V.~K. Nagaraja, V.~I. Morariu, and L.~S. Davis, ``Modeling context between objects for referring expression understanding,'' in \emph{Computer Vision--ECCV 2016: 14th European Conference, Amsterdam, The Netherlands, October 11--14, 2016, Proceedings, Part IV 14}.\hskip 1em plus 0.5em minus 0.4em\relax Springer, 2016, pp. 792--807.

\bibitem{wang2024visionllm}
W.~Wang, Z.~Chen, X.~Chen, J.~Wu, X.~Zhu, G.~Zeng, P.~Luo, T.~Lu, J.~Zhou, Y.~Qiao \emph{et~al.}, ``Visionllm: Large language model is also an open-ended decoder for vision-centric tasks,'' \emph{Advances in Neural Information Processing Systems}, vol.~36, 2024.

\bibitem{zhao2023bubogpt}
Y.~Zhao, Z.~Lin, D.~Zhou, Z.~Huang, J.~Feng, and B.~Kang, ``Bubogpt: Enabling visual grounding in multi-modal llms,'' \emph{arXiv preprint arXiv:2307.08581}, 2023.

\bibitem{ma2024groma}
C.~Ma, Y.~Jiang, J.~Wu, Z.~Yuan, and X.~Qi, ``Groma: Localized visual tokenization for grounding multimodal large language models,'' \emph{arXiv preprint arXiv:2404.13013}, 2024.

\bibitem{lai2024lisa}
X.~Lai, Z.~Tian, Y.~Chen, Y.~Li, Y.~Yuan, S.~Liu, and J.~Jia, ``Lisa: Reasoning segmentation via large language model,'' in \emph{Proceedings of the IEEE/CVF Conference on Computer Vision and Pattern Recognition}, 2024, pp. 9579--9589.

\bibitem{rasheed2024glamm}
H.~Rasheed, M.~Maaz, S.~Shaji, A.~Shaker, S.~Khan, H.~Cholakkal, R.~M. Anwer, E.~Xing, M.-H. Yang, and F.~S. Khan, ``Glamm: Pixel grounding large multimodal model,'' in \emph{Proceedings of the IEEE/CVF Conference on Computer Vision and Pattern Recognition}, 2024, pp. 13\,009--13\,018.

\bibitem{zhang2023nextchat}
A.~Zhang, L.~Zhao, C.-W. Xie, Y.~Zheng, W.~Ji, and T.-S. Chua, ``Next-chat: An lmm for chat, detection and segmentation,'' \emph{arXiv preprint arXiv:2311.04498}, 2023.

\bibitem{GREC}
S.~He, H.~Ding, C.~Liu, and X.~Jiang, ``{GREC}: Generalized referring expression comprehension,'' \emph{arXiv preprint arXiv:2308.16182}, 2023.

\bibitem{flickr30k}
\BIBentryALTinterwordspacing
B.~A. Plummer, L.~Wang, C.~M. Cervantes, J.~C. Caicedo, J.~Hockenmaier, and S.~Lazebnik, ``\BIBforeignlanguage{en-US}{Flickr30k entities: Collecting region-to-phrase correspondences for richer image-to-sentence models},'' \emph{\BIBforeignlanguage{en-US}{International Journal of Computer Vision}}, p. 74–93, May 2017. [Online]. Available: \url{http://dx.doi.org/10.1007/s11263-016-0965-7}
\BIBentrySTDinterwordspacing

\bibitem{visualgenemo}
\BIBentryALTinterwordspacing
R.~Krishna, Y.~Zhu, O.~Groth, J.~Johnson, K.~Hata, J.~Kravitz, S.~Chen, Y.~Kalantidis, L.-J. Li, D.~A. Shamma, M.~S. Bernstein, and L.~Fei-Fei, ``\BIBforeignlanguage{en-US}{Visual genome: Connecting language and vision using crowdsourced dense image annotations},'' \emph{\BIBforeignlanguage{en-US}{International Journal of Computer Vision}}, p. 32–73, May 2017. [Online]. Available: \url{http://dx.doi.org/10.1007/s11263-016-0981-7}
\BIBentrySTDinterwordspacing

\bibitem{objects}
S.~Shao, Z.~Li, T.~Zhang, C.~Peng, G.~Yu, X.~Zhang, J.~Li, and J.~Sun, ``Objects365: A large-scale, high-quality dataset for object detection,'' in \emph{2019 IEEE/CVF International Conference on Computer Vision (ICCV)}, 2019, pp. 8429--8438.

\bibitem{lin2014microsoft}
T.-Y. Lin, M.~Maire, S.~Belongie, J.~Hays, P.~Perona, D.~Ramanan, P.~Doll{\'a}r, and C.~L. Zitnick, ``Microsoft coco: Common objects in context,'' in \emph{ECCV}, 2014.

\bibitem{wang2023v3det}
J.~Wang, P.~Zhang, T.~Chu, Y.~Cao, Y.~Zhou, T.~Wu, B.~Wang, C.~He, and D.~Lin, ``V3det: Vast vocabulary visual detection dataset,'' in \emph{The IEEE International Conference on Computer Vision (ICCV)}, October 2023.

\bibitem{spaCy}
M.~Honnibal, I.~Montani, S.~Van~Landeghem, and A.~Boyd, ``{spaCy: Industrial-strength Natural Language Processing in Python},'' 2020.

\bibitem{ding2023enhancing}
N.~Ding, Y.~Chen, B.~Xu, Y.~Qin, Z.~Zheng, S.~Hu, Z.~Liu, M.~Sun, and B.~Zhou, ``Enhancing chat language models by scaling high-quality instructional conversations,'' \emph{arXiv preprint arXiv:2305.14233}, 2023.

\bibitem{ghosal2023flacuna}
D.~Ghosal, Y.~K. Chia, N.~Majumder, and S.~Poria, ``Flacuna: Unleashing the problem solving power of vicuna using flan fine-tuning,'' 2023.

\bibitem{OpenOrca}
W.~Lian, B.~Goodson, E.~Pentland, A.~Cook, C.~Vong, and "Teknium", ``Openorca: An open dataset of gpt augmented flan reasoning traces,'' \url{https://https://huggingface.co/Open-Orca/OpenOrca}, 2023.

\bibitem{yu2023metamath}
L.~Yu, W.~Jiang, H.~Shi, J.~Yu, Z.~Liu, Y.~Zhang, J.~T. Kwok, Z.~Li, A.~Weller, and W.~Liu, ``Metamath: Bootstrap your own mathematical questions for large language models,'' \emph{arXiv preprint arXiv:2309.12284}, 2023.

\bibitem{yue2023mammoth}
X.~Yue, X.~Qu, G.~Zhang, Y.~Fu, W.~Huang, H.~Sun, Y.~Su, and W.~Chen, ``Mammoth: Building math generalist models through hybrid instruction tuning,'' \emph{arXiv preprint arXiv:2309.05653}, 2023.

\bibitem{luo2023wizardcoder}
Z.~Luo, C.~Xu, P.~Zhao, Q.~Sun, X.~Geng, W.~Hu, C.~Tao, J.~Ma, Q.~Lin, and D.~Jiang, ``Wizardcoder: Empowering code large language models with evol-instruct,'' \emph{arXiv preprint arXiv:2306.08568}, 2023.

\bibitem{chen2024allava}
G.~H. Chen, S.~Chen, R.~Zhang, J.~Chen, X.~Wu, Z.~Zhang, Z.~Chen, J.~Li, X.~Wan, and B.~Wang, ``Allava: Harnessing gpt4v-synthesized data for a lite vision-language model,'' 2024.

\bibitem{wang2023instruct4v}
J.~Wang, L.~Meng, Z.~Weng, B.~He, Z.~Wu, and Y.-G. Jiang, ``To see is to believe: Prompting gpt-4v for better visual instruction tuning,'' \emph{arXiv preprint arXiv:2311.07574}, 2023.

\bibitem{sidorov2020textcaps}
O.~Sidorov, R.~Hu, M.~Rohrbach, and A.~Singh, ``Textcaps: a dataset for image captioning with reading comprehension,'' in \emph{Computer Vision--ECCV 2020: 16th European Conference, Glasgow, UK, August 23--28, 2020, Proceedings, Part II 16}.\hskip 1em plus 0.5em minus 0.4em\relax Springer, 2020, pp. 742--758.

\bibitem{vqa}
\BIBentryALTinterwordspacing
Y.~Goyal, T.~Khot, A.~Agrawal, D.~Summers-Stay, D.~Batra, and D.~Parikh, ``\BIBforeignlanguage{en-US}{Making the v in vqa matter: Elevating the role of image understanding in visual question answering},'' \emph{\BIBforeignlanguage{en-US}{International Journal of Computer Vision}}, p. 398–414, Apr 2019. [Online]. Available: \url{http://dx.doi.org/10.1007/s11263-018-1116-0}
\BIBentrySTDinterwordspacing

\bibitem{gqa}
\BIBentryALTinterwordspacing
D.~A. Hudson and C.~D. Manning, ``\BIBforeignlanguage{en-US}{Gqa: A new dataset for real-world visual reasoning and compositional question answering},'' in \emph{\BIBforeignlanguage{en-US}{2019 IEEE/CVF Conference on Computer Vision and Pattern Recognition (CVPR)}}, Jun 2019. [Online]. Available: \url{http://dx.doi.org/10.1109/cvpr.2019.00686}
\BIBentrySTDinterwordspacing

\bibitem{marino2019ok}
K.~Marino, M.~Rastegari, A.~Farhadi, and R.~Mottaghi, ``Ok-vqa: A visual question answering benchmark requiring external knowledge,'' in \emph{Proceedings of the IEEE/cvf conference on computer vision and pattern recognition}, 2019, pp. 3195--3204.

\bibitem{schwenk2022okvqa}
D.~Schwenk, A.~Khandelwal, C.~Clark, K.~Marino, and R.~Mottaghi, ``A-okvqa: A benchmark for visual question answering using world knowledge,'' in \emph{European conference on computer vision}.\hskip 1em plus 0.5em minus 0.4em\relax Springer, 2022, pp. 146--162.

\bibitem{sqa}
P.~Lu, S.~Mishra, T.~Xia, L.~Qiu, K.-W. Chang, S.-C. Zhu, O.~Tafjord, P.~Clark, and A.~Kalyan, ``Learn to explain: Multimodal reasoning via thought chains for science question answering,'' in \emph{The 36th Conference on Neural Information Processing Systems (NeurIPS)}, 2022.

\bibitem{bigham2010vizwiz}
J.~P. Bigham, C.~Jayant, H.~Ji, G.~Little, A.~Miller, R.~C. Miller, R.~Miller, A.~Tatarowicz, B.~White, S.~White \emph{et~al.}, ``Vizwiz: nearly real-time answers to visual questions,'' in \emph{Proceedings of the 23nd annual ACM symposium on User interface software and technology}, 2010, pp. 333--342.

\bibitem{textvqa}
A.~Singh, V.~Natarajan, M.~Shah, Y.~Jiang, X.~Chen, D.~Batra, D.~Parikh, and M.~Rohrbach, ``Towards vqa models that can read,'' in \emph{Proceedings of the IEEE/CVF conference on computer vision and pattern recognition}, 2019, pp. 8317--8326.

\bibitem{ocrvqa}
A.~Mishra, S.~Shekhar, A.~K. Singh, and A.~Chakraborty, ``Ocr-vqa: Visual question answering by reading text in images,'' in \emph{ICDAR}, 2019.

\bibitem{ai2d}
A.~Kembhavi, M.~Salvato, E.~Kolve, M.~Seo, H.~Hajishirzi, and A.~Farhadi, ``A diagram is worth a dozen images,'' in \emph{Computer Vision--ECCV 2016: 14th European Conference, Amsterdam, The Netherlands, October 11--14, 2016, Proceedings, Part IV 14}.\hskip 1em plus 0.5em minus 0.4em\relax Springer, 2016, pp. 235--251.

\bibitem{kim2022donut}
G.~Kim, T.~Hong, M.~Yim, J.~Nam, J.~Park, J.~Yim, W.~Hwang, S.~Yun, D.~Han, and S.~Park, ``Ocr-free document understanding transformer,'' in \emph{European Conference on Computer Vision (ECCV)}, 2022.

\bibitem{kafle2018dvqa}
K.~Kafle, S.~Cohen, B.~Price, and C.~Kanan, ``Dvqa: Understanding data visualizations via question answering,'' in \emph{CVPR}, 2018.

\bibitem{masry-etal-2022-chartqa}
\BIBentryALTinterwordspacing
A.~Masry, D.~Long, J.~Q. Tan, S.~Joty, and E.~Hoque, ``{C}hart{QA}: A benchmark for question answering about charts with visual and logical reasoning,'' in \emph{Findings of the Association for Computational Linguistics: ACL 2022}.\hskip 1em plus 0.5em minus 0.4em\relax Dublin, Ireland: Association for Computational Linguistics, May 2022, pp. 2263--2279. [Online]. Available: \url{https://aclanthology.org/2022.findings-acl.177}
\BIBentrySTDinterwordspacing

\bibitem{mathew2020docvqa}
M.~Mathew, D.~Karatzas, R.~Manmatha, and C.~Jawahar, ``Docvqa: a dataset for vqa on document images. corr abs/2007.00398 (2020),'' \emph{arXiv preprint arXiv:2007.00398}, 2020.

\bibitem{mathew2022infographicvqa}
M.~Mathew, V.~Bagal, R.~Tito, D.~Karatzas, E.~Valveny, and C.~Jawahar, ``Infographicvqa,'' in \emph{Proceedings of the IEEE/CVF Winter Conference on Applications of Computer Vision}, 2022, pp. 1697--1706.

\bibitem{borchmann2021due}
{\L}.~Borchmann, M.~Pietruszka, T.~Stanislawek, D.~Jurkiewicz, M.~Turski, K.~Szyndler, and F.~Grali{\'n}ski, ``Due: End-to-end document understanding benchmark,'' in \emph{Thirty-fifth Conference on Neural Information Processing Systems Datasets and Benchmarks Track (Round 2)}, 2021.

\bibitem{OpenImages}
A.~Kuznetsova, H.~Rom, N.~Alldrin, J.~Uijlings, I.~Krasin, J.~Pont-Tuset, S.~Kamali, S.~Popov, M.~Malloci, T.~Duerig, and V.~Ferrari, ``The open images dataset v4: Unified image classification, object detection, and visual relationship detection at scale,'' \emph{arXiv:1811.00982}, 2018.

\bibitem{fscd}
K.~N. Thanh~Nguyen, Chau~Pham and M.~Hoai, ``{Few-shot Object Counting and Detection},'' in \emph{Proceedings of the European Conference on Computer Vision 2022}, 2022.

\bibitem{huitt2003piaget}
W.~Huitt and J.~Hummel, ``Piaget's theory of cognitive development,'' \emph{Educational psychology interactive}, vol.~3, no.~2, pp. 1--5, 2003.

\bibitem{bengio2009curriculum}
Y.~Bengio, J.~Louradour, R.~Collobert, and J.~Weston, ``Curriculum learning,'' in \emph{Proceedings of the 26th annual international conference on machine learning}, 2009, pp. 41--48.

\bibitem{radford2021clip}
A.~Radford, J.~W. Kim, C.~Hallacy, A.~Ramesh, G.~Goh, S.~Agarwal, G.~Sastry, A.~Askell, P.~Mishkin, J.~Clark \emph{et~al.}, ``Learning transferable visual models from natural language supervision,'' in \emph{International conference on machine learning}.\hskip 1em plus 0.5em minus 0.4em\relax PMLR, 2021, pp. 8748--8763.

\bibitem{team2024gemma}
G.~Team, M.~Riviere, S.~Pathak, P.~G. Sessa, C.~Hardin, S.~Bhupatiraju, L.~Hussenot, T.~Mesnard, B.~Shahriari, A.~Ram{\'e} \emph{et~al.}, ``Gemma 2: Improving open language models at a practical size,'' \emph{arXiv preprint arXiv:2408.00118}, 2024.

\bibitem{adam}
I.~Loshchilov and F.~Hutter, ``Decoupled weight decay regularization,'' \emph{arXiv preprint arXiv:1711.05101}, 2017.

\bibitem{rasley2020deepspeed}
J.~Rasley, S.~Rajbhandari, O.~Ruwase, and Y.~He, ``Deepspeed: System optimizations enable training deep learning models with over 100 billion parameters,'' in \emph{Proceedings of the 26th ACM SIGKDD International Conference on Knowledge Discovery \& Data Mining}, 2020, pp. 3505--3506.

\bibitem{cosine}
I.~Loshchilov and F.~Hutter, ``Sgdr: Stochastic gradient descent with warm restarts,'' \emph{arXiv preprint arXiv:1608.03983}, 2016.

\bibitem{warmup}
P.~Goyal, P.~Doll{\'a}r, R.~Girshick, P.~Noordhuis, L.~Wesolowski, A.~Kyrola, A.~Tulloch, Y.~Jia, and K.~He, ``Accurate, large minibatch sgd: Training imagenet in 1 hour,'' \emph{arXiv preprint arXiv:1706.02677}, 2017.

\bibitem{ye2024mplug}
Q.~Ye, H.~Xu, J.~Ye, M.~Yan, A.~Hu, H.~Liu, Q.~Qian, J.~Zhang, and F.~Huang, ``mplug-owl2: Revolutionizing multi-modal large language model with modality collaboration,'' in \emph{Proceedings of the IEEE/CVF Conference on Computer Vision and Pattern Recognition}, 2024, pp. 13\,040--13\,051.

\bibitem{you2024ferret}
K.~You, H.~Zhang, E.~Schoop, F.~Weers, A.~Swearngin, J.~Nichols, Y.~Yang, and Z.~Gan, ``Ferret-ui: Grounded mobile ui understanding with multimodal llms,'' \emph{arXiv preprint arXiv:2404.05719}, 2024.

\bibitem{li2023otterhd}
B.~Li, P.~Zhang, J.~Yang, Y.~Zhang, F.~Pu, and Z.~Liu, ``Otterhd: A high-resolution multi-modality model,'' \emph{arXiv preprint arXiv:2311.04219}, 2023.

\bibitem{zhang2024omgllava}
T.~Zhang, X.~Li, H.~Fei, H.~Yuan, S.~Wu, S.~Ji, C.~C. Loy, and S.~Yan, ``Omg-llava: Bridging image-level, object-level, pixel-level reasoning and understanding,'' in \emph{The 38th Conference on Neural Information Processing Systems (NeurIPS)}, 2024.

\bibitem{pope}
Y.~Li, Y.~Du, K.~Zhou, J.~Wang, W.~X. Zhao, and J.-R. Wen, ``Evaluating object hallucination in large vision-language models,'' \emph{arXiv preprint arXiv:2305.10355}, 2023.

\bibitem{fu2023mme}
C.~Fu, P.~Chen, Y.~Shen, Y.~Qin, M.~Zhang, X.~Lin, Z.~Qiu, W.~Lin, J.~Yang, X.~Zheng \emph{et~al.}, ``Mme: a comprehensive evaluation benchmark for multimodal large language models. corr abs/2306.13394 (2023),'' 2023.

\bibitem{MMBench}
Y.~Liu, H.~Duan, Y.~Zhang, B.~Li, S.~Zhang, W.~Zhao, Y.~Yuan, J.~Wang, C.~He, Z.~Liu \emph{et~al.}, ``Mmbench: Is your multi-modal model an all-around player?'' \emph{arXiv preprint arXiv:2307.06281}, 2023.

\bibitem{yue2023mmmu}
X.~Yue, Y.~Ni, K.~Zhang, T.~Zheng, R.~Liu, G.~Zhang, S.~Stevens, D.~Jiang, W.~Ren, Y.~Sun \emph{et~al.}, ``Mmmu: A massive multi-discipline multimodal understanding and reasoning benchmark for expert agi. arxiv,'' 2023.

\bibitem{lee2023pix2struct}
K.~Lee, M.~Joshi, I.~R. Turc, H.~Hu, F.~Liu, J.~M. Eisenschlos, U.~Khandelwal, P.~Shaw, M.-W. Chang, and K.~Toutanova, ``Pix2struct: Screenshot parsing as pretraining for visual language understanding,'' in \emph{International Conference on Machine Learning}.\hskip 1em plus 0.5em minus 0.4em\relax PMLR, 2023, pp. 18\,893--18\,912.

\bibitem{ye2023ureader}
J.~Ye, A.~Hu, H.~Xu, Q.~Ye, M.~Yan, G.~Xu, C.~Li, J.~Tian, Q.~Qian, J.~Zhang \emph{et~al.}, ``Ureader: Universal ocr-free visually-situated language understanding with multimodal large language model,'' \emph{arXiv preprint arXiv:2310.05126}, 2023.

\bibitem{hu2024mplug}
A.~Hu, H.~Xu, J.~Ye, M.~Yan, L.~Zhang, B.~Zhang, C.~Li, J.~Zhang, Q.~Jin, F.~Huang \emph{et~al.}, ``mplug-docowl 1.5: Unified structure learning for ocr-free document understanding,'' \emph{arXiv preprint arXiv:2403.12895}, 2024.

\bibitem{liu2024hrvda}
C.~Liu, K.~Yin, H.~Cao, X.~Jiang, X.~Li, Y.~Liu, D.~Jiang, X.~Sun, and L.~Xu, ``Hrvda: High-resolution visual document assistant,'' in \emph{Proceedings of the IEEE/CVF Conference on Computer Vision and Pattern Recognition}, 2024, pp. 15\,534--15\,545.

\bibitem{ren2015faster}
S.~Ren, K.~He, R.~Girshick, and J.~Sun, ``Faster r-cnn: Towards real-time object detection with region proposal networks,'' \emph{Advances in neural information processing systems}, vol.~28, 2015.

\bibitem{lin2017feature}
T.-Y. Lin, P.~Doll{\'a}r, R.~Girshick, K.~He, B.~Hariharan, and S.~Belongie, ``Feature pyramid networks for object detection,'' in \emph{Proceedings of the IEEE conference on computer vision and pattern recognition}, 2017, pp. 2117--2125.

\bibitem{liu2022dab}
S.~Liu, F.~Li, H.~Zhang, X.~Yang, X.~Qi, H.~Su, J.~Zhu, and L.~Zhang, ``Dab-detr: Dynamic anchor boxes are better queries for detr,'' \emph{arXiv preprint arXiv:2201.12329}, 2022.

\bibitem{chen2021pix2seq}
T.~Chen, S.~Saxena, L.~Li, D.~J. Fleet, and G.~Hinton, ``Pix2seq: A language modeling framework for object detection,'' \emph{arXiv preprint arXiv:2109.10852}, 2021.

\bibitem{carion2020end}
N.~Carion, F.~Massa, G.~Synnaeve, N.~Usunier, A.~Kirillov, and S.~Zagoruyko, ``End-to-end object detection with transformers,'' in \emph{European conference on computer vision}.\hskip 1em plus 0.5em minus 0.4em\relax Springer, 2020, pp. 213--229.

\bibitem{kamath2021mdetr}
A.~Kamath, M.~Singh, Y.~LeCun, G.~Synnaeve, I.~Misra, and N.~Carion, ``Mdetr-modulated detection for end-to-end multi-modal understanding,'' in \emph{Proceedings of the IEEE/CVF international conference on computer vision}, 2021, pp. 1780--1790.

\bibitem{liu2023grounding}
S.~Liu, Z.~Zeng, T.~Ren, F.~Li, H.~Zhang, J.~Yang, C.~Li, J.~Yang, H.~Su, J.~Zhu \emph{et~al.}, ``Grounding dino: Marrying dino with grounded pre-training for open-set object detection,'' \emph{arXiv preprint arXiv:2303.05499}, 2023.

\bibitem{wang2022ofa}
P.~Wang, A.~Yang, R.~Men, J.~Lin, S.~Bai, Z.~Li, J.~Ma, C.~Zhou, J.~Zhou, and H.~Yang, ``Ofa: Unifying architectures, tasks, and modalities through a simple sequence-to-sequence learning framework,'' in \emph{International conference on machine learning}.\hskip 1em plus 0.5em minus 0.4em\relax PMLR, 2022, pp. 23\,318--23\,340.

\bibitem{pramanick2024jack}
S.~Pramanick, G.~Han, R.~Hou, S.~Nag, S.-N. Lim, N.~Ballas, Q.~Wang, R.~Chellappa, and A.~Almahairi, ``Jack of all tasks master of many: Designing general-purpose coarse-to-fine vision-language model,'' in \emph{Proceedings of the IEEE/CVF Conference on Computer Vision and Pattern Recognition}, 2024, pp. 14\,076--14\,088.

\bibitem{kirillov2023segany}
A.~Kirillov, E.~Mintun, N.~Ravi, H.~Mao, C.~Rolland, L.~Gustafson, T.~Xiao, S.~Whitehead, A.~C. Berg, W.-Y. Lo, P.~Doll{\'a}r, and R.~Girshick, ``Segment anything,'' \emph{arXiv:2304.02643}, 2023.

\bibitem{oquab2023dinov2}
M.~Oquab, T.~Darcet, T.~Moutakanni, H.~V. Vo, M.~Szafraniec, V.~Khalidov, P.~Fernandez, D.~Haziza, F.~Massa, A.~El-Nouby, R.~Howes, P.-Y. Huang, H.~Xu, V.~Sharma, S.-W. Li, W.~Galuba, M.~Rabbat, M.~Assran, N.~Ballas, G.~Synnaeve, I.~Misra, H.~Jegou, J.~Mairal, P.~Labatut, A.~Joulin, and P.~Bojanowski, ``Dinov2: Learning robust visual features without supervision,'' 2023.

\end{thebibliography}

\newpage

\section{Biography Section}
 

\begin{IEEEbiography}[{\includegraphics[width=1in,height=1.25in,clip,keepaspectratio]{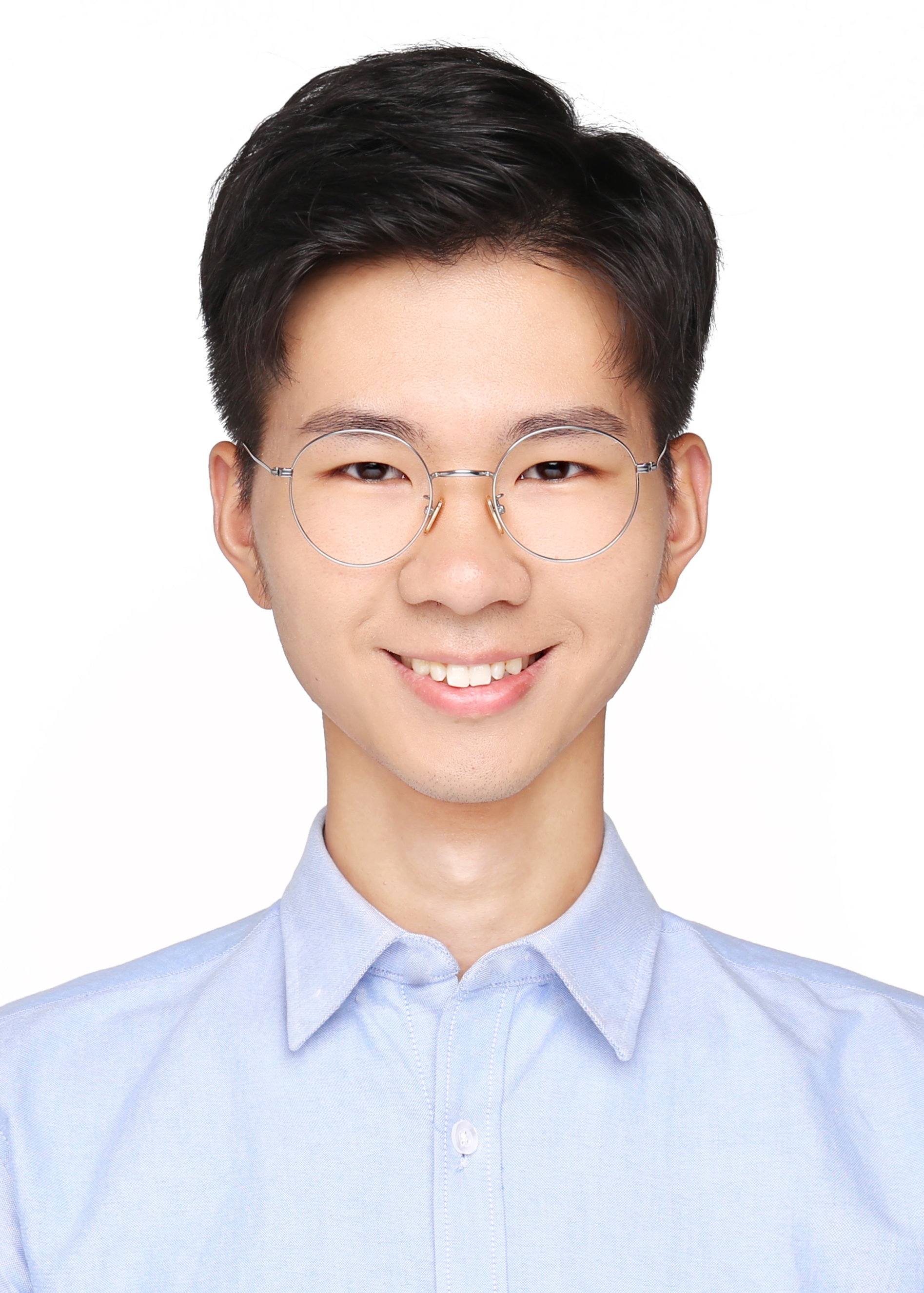}}]{Yufei Zhan}
received his B.E. degree from the University of Science and Technology Beijing, Beijing, China, in 2021. He is currently pursuing the Ph.D. degree in Pattern Recognition and Intelligent Systems at the Foundation Model Research Center, Institute of Automation, Chinese Academy of Sciences. His research interests include large multimodal models, visual reasoning, open vocabulary object detection, and computer vision.
\end{IEEEbiography}

\vspace{11pt}

\begin{IEEEbiography}[{\includegraphics[width=1in,height=1.25in,clip,keepaspectratio]{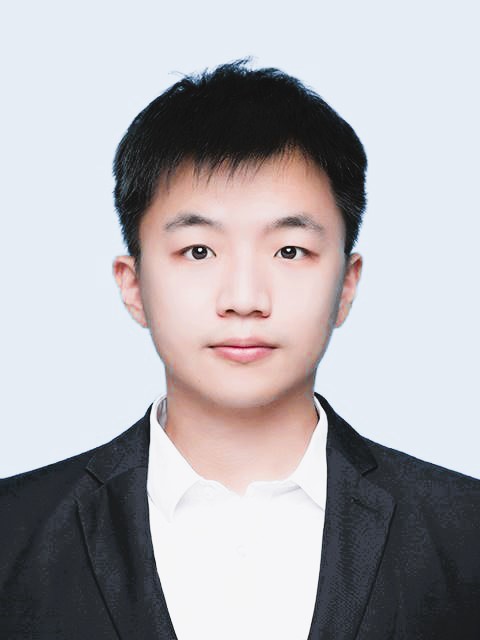}}]{Hongyin Zhao}
received his Bachelor's degree from Southeast University in Nanjing, Jiangsu Province, in 2020, and obtained his Master's degree in Artificial Intelligence from the University of Chinese Academy of Sciences in Beijing in 2023. He is currently an engineer at the Institute of Automation, Chinese Academy of Sciences. His main research interests include object detection and large multimodal models.
\end{IEEEbiography}

\vspace{11pt}

\begin{IEEEbiography}[{\includegraphics[width=1in,height=1.25in,clip,keepaspectratio]{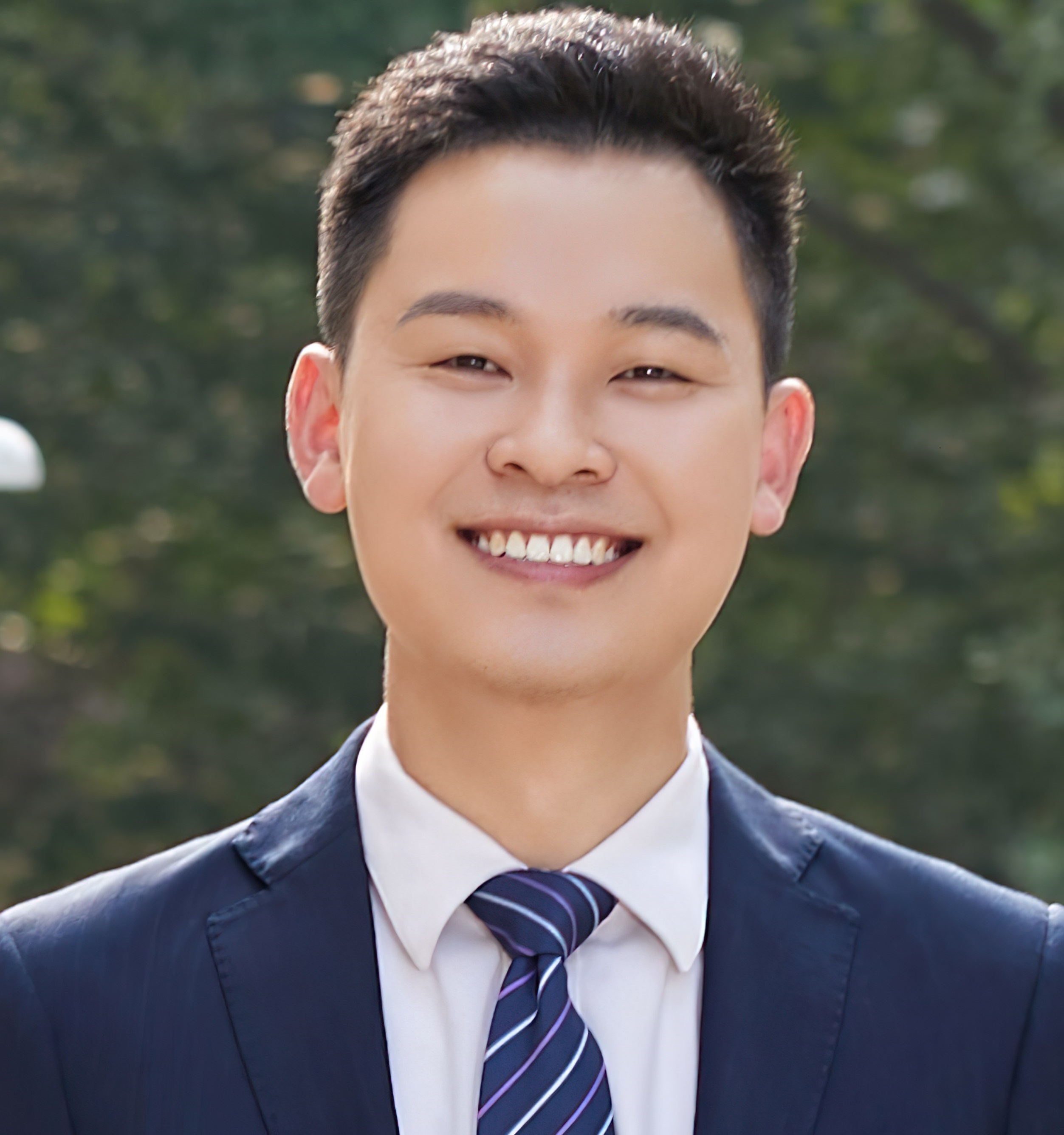}}]{Yousong Zhu}
received the B.E. degree in 2014 from Central South University, Changsha, China and the Ph.D. degree in pattern recognition and intelligence systems from the Institute of Automation, Chinese Academy of Sciences, in 2019. He is currently an Associate Professor with the Foundation Model Research Center, Institute of Automation, Chinese Academy of Sciences. His current research interests include visual-language model, computer vision and object detection, and self- supervised learning and general vision model.
\end{IEEEbiography}

\vspace{11pt}

\begin{IEEEbiography}[{\includegraphics[width=1in,height=1.25in,clip,keepaspectratio]{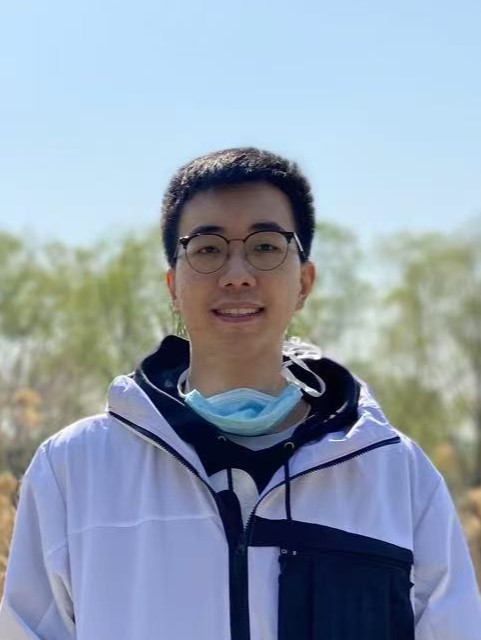}}]{Fan Yang}
received the B.E. degree in Hebei University in China in 2016 and the M.S. degrees from the Beihang University of China in 2020. He worked as an algorithm researcher at Sensetime Technology Co., Ltd Beijing, China, in 2022. He is currently pursuing the Ph.D. degree at the Institute of Automation, Chinese Academy of Sciences. His research interests include multi-model downstream task learning in computer vision.
\end{IEEEbiography}

\vspace{11pt}

\begin{IEEEbiography}[{\includegraphics[width=1in,height=1.25in,clip,keepaspectratio]{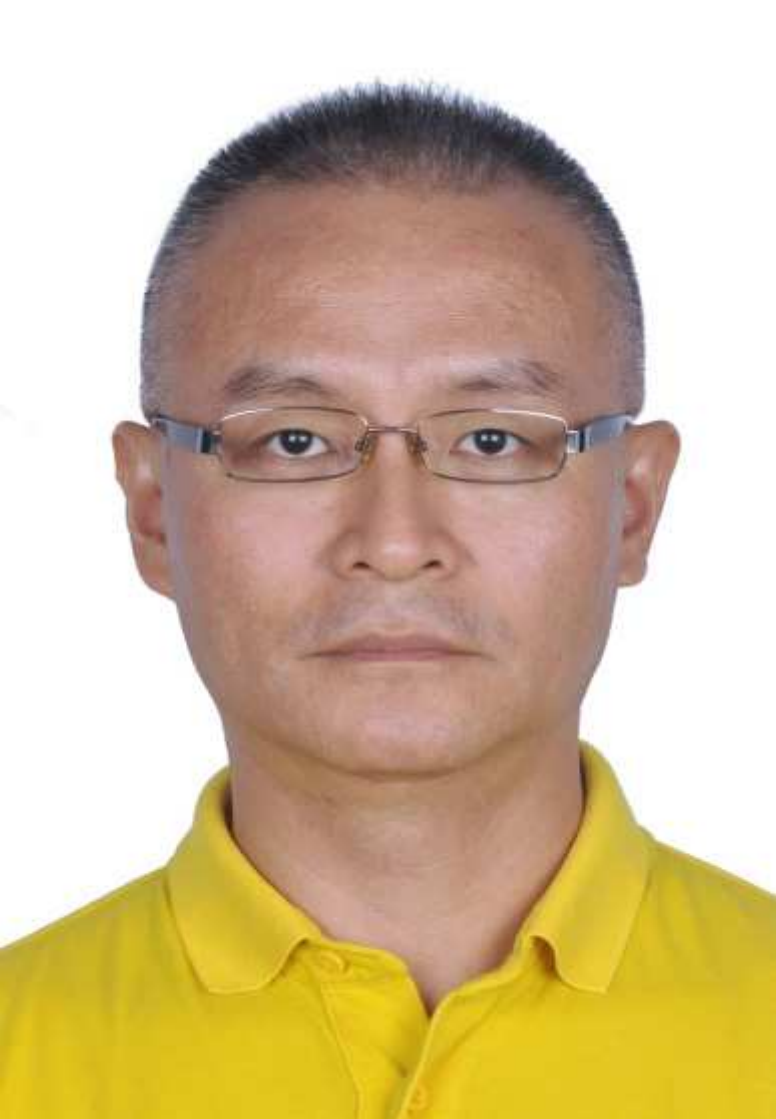}}]{Ming Tang}
received the B.S. degree in computer science and engineering and M.S. degree in artificial intelligence from Zhejiang University, Hangzhou, China, in 1984 and 1987, respectively, and the Ph.D. degree in pattern recognition and intelligent system from the Chinese Academy of Sciences, Beijing, China, in 2002. He is currently a Professor with the Foundation Model Research Center, Institute of Automation, Chinese Academy of Sciences. His current research interests include computer vision and machine learning.
\end{IEEEbiography}

\vspace{11pt}

\begin{IEEEbiography}[{\includegraphics[width=1in,height=1.25in,clip,keepaspectratio]{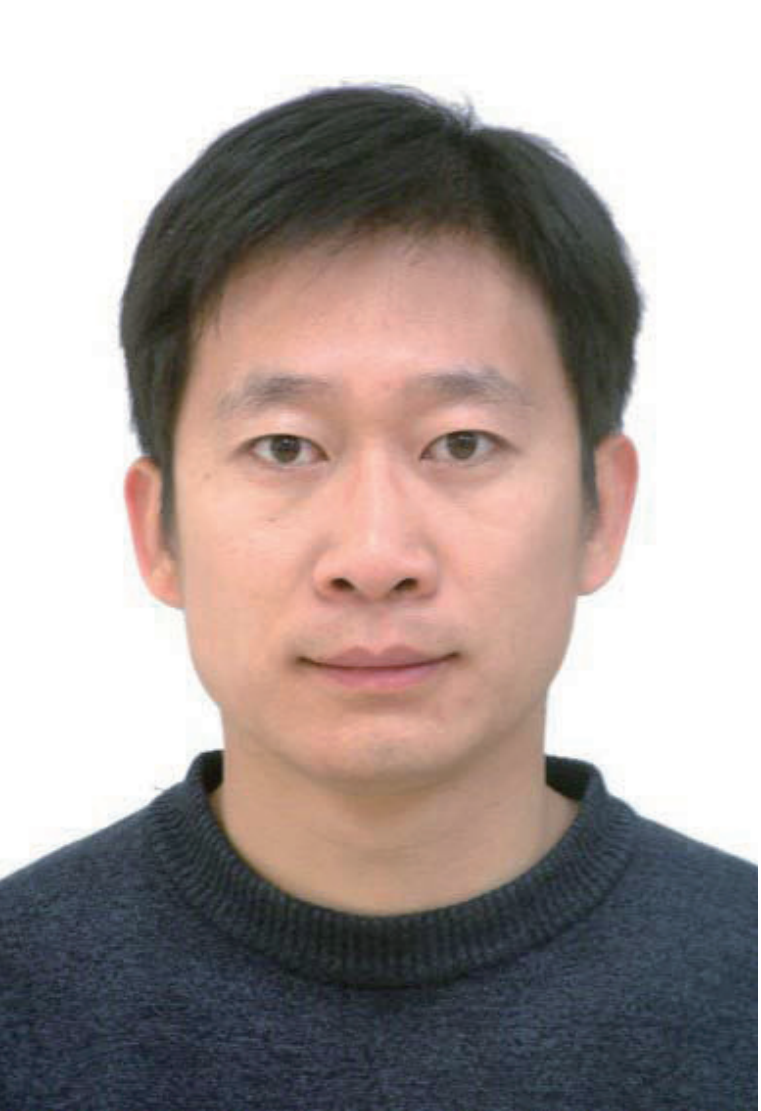}}]{Jinqiao Wang}
received the B.E. degree from the Hebei University of Technology, China, in 2001, the M.S. degree from Tianjin University, China, in 2004, and the Ph.D. degree in pattern recognition and intelligence systems from the Foundation Model Research Center, Chinese Academy of Sciences, in 2008. He is currently a Professor with the Chinese Academy of Sciences. His research interests include pattern recognition and machine learning, image and video processing, mobile multimedia, and intelligent video surveillance.
\end{IEEEbiography}
\vfill

\end{document}